\documentclass[10pt,twocolumn,letterpaper]{article}

\usepackage{cvpr}              

\usepackage{graphicx}
\usepackage{amsmath}
\usepackage{amssymb}
\usepackage{booktabs}
\usepackage{multirow}
\usepackage{tabularx}
\usepackage{color}
\usepackage{enumerate}
\usepackage{enumitem}
\usepackage[misc]{ifsym}
\usepackage{bbding}
\usepackage[pagebackref,breaklinks,colorlinks]{hyperref}

\usepackage[capitalize]{cleveref}
\crefname{section}{Sec.}{Secs.}
\Crefname{section}{Section}{Sections}
\Crefname{table}{Table}{Tables}
\crefname{table}{Tab.}{Tabs.}

\def\cvprPaperID{xxxx} 
\def\confName{CVPR}
\def\confYear{2022}

\begin{document}

\title{Label Matching Semi-Supervised Object Detection}

\author{Binbin Chen$^2$, Weijie Chen$^{1,2,}$\footnotemark[2], Shicai Yang$^2$, Yunyi Xuan$^2$, Jie Song$^1$ \\
Di Xie$^2$, Shiliang Pu$^2$, Mingli Song$^1$, Yueting Zhuang$^{1,}$\footnotemark[2]\\
{\normalsize $^1$Zhejiang University, $^2$Hikvision Research Institute}\\
{\tt\small \{chenbinbin8,chenweijie5,yangshicai,xuanyunyi,xiedi,pushiliang.hri\}@hikvision.com} \\
{\tt\small \{sjie,songml,yzhuang\}@zju.edu.cn}
}
\maketitle
\renewcommand{\thefootnote}{\fnsymbol{footnote}}
\footnotetext[2]{Corresponding author}

\begin{abstract}
Semi-supervised object detection has made significant progress with the development of mean teacher driven self-training. Despite the promising results, the label mismatch problem is not yet fully explored in the previous works, leading to severe confirmation bias during self-training. In this paper, we delve into this problem and propose a simple yet effective LabelMatch framework from two different yet complementary perspectives, i.e., distribution-level and instance-level. For the former one, it is reasonable to approximate the class distribution of the unlabeled data from that of the labeled data according to Monte Carlo Sampling. Guided by this weakly supervision cue, we introduce a re-distribution mean teacher, which leverages adaptive label-distribution-aware confidence thresholds to generate unbiased pseudo labels to drive student learning. For the latter one, there exists an overlooked label assignment ambiguity problem across teacher-student models. To remedy this issue, we present a novel label assignment mechanism for self-training framework, namely proposal self-assignment, which injects the proposals from student into teacher and generates accurate pseudo labels to match each proposal in the student model accordingly. Experiments on both MS-COCO and PASCAL-VOC datasets demonstrate the considerable superiority of our proposed framework to other state-of-the-arts. Code will be available at \url{https://github.com/hikvision-research/SSOD}.

\end{abstract}

\section{Introduction}
\label{sec:intro}

\begin{figure}
  \centering
  \includegraphics[width=\linewidth]{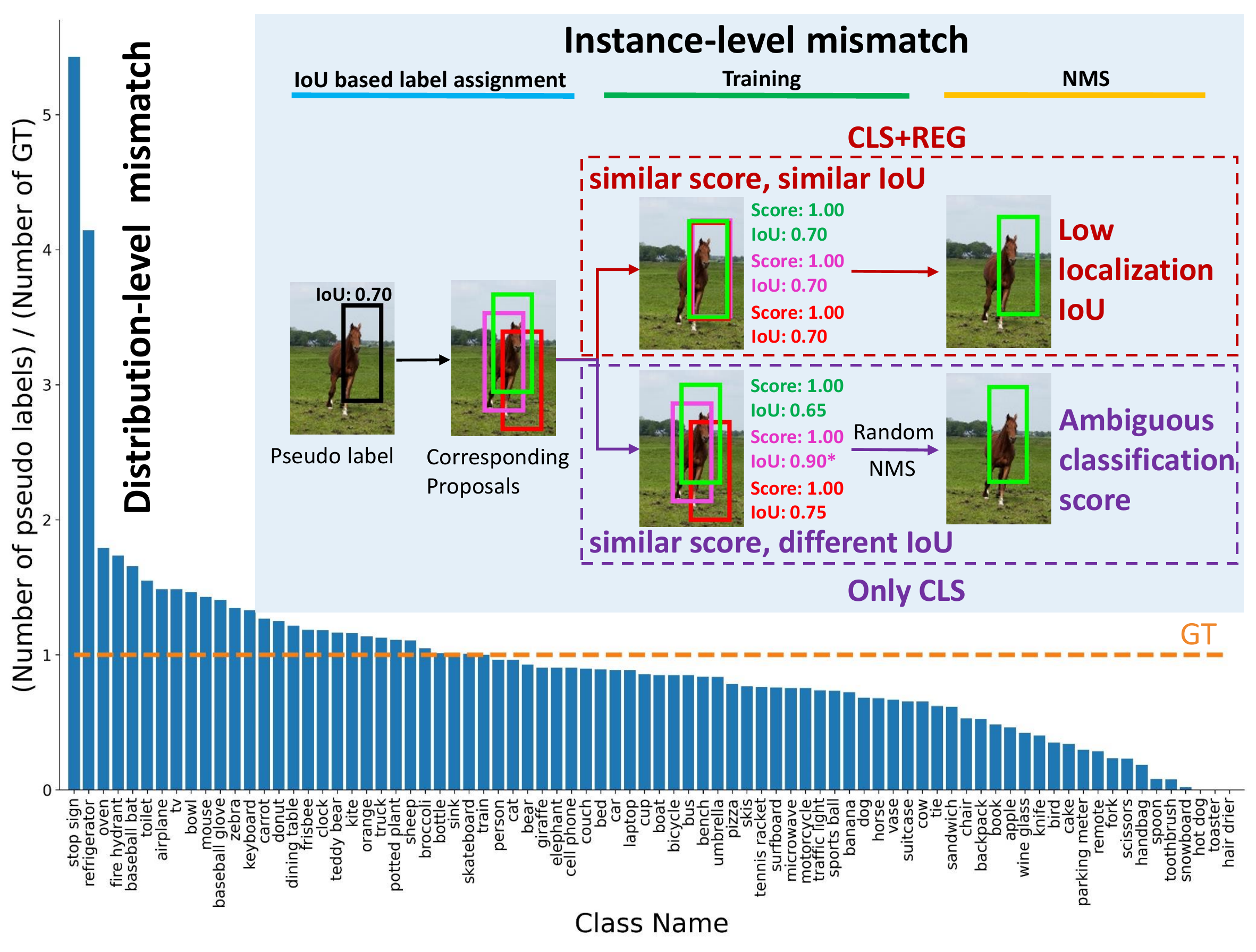}
  \vskip -0.10in
  \caption{Label mismatch problems on the MS-COCO dataset. 1) \textbf{Distribution-level mismatch}: there exists a bias between the pseudo labels produced by the single confidence threshold and the ground truth labels (GT) during self-training, as shown in the relation of the blue bar and the orange dotted line. 2) \textbf{Instance-level mismatch}: there are two kinds of training patterns for the unlabeled data in the previous SSOD frameworks. One is the same as supervised learning, using both classification and box regression for optimization, which will overfit the poor-quality pseudo labels and result in low localization accuracy. To avoid incorrect box regression, another one merely exploits a classification objective \cite{liu2021unbiased}, which will bring ambiguity due to the similar classification scores to confuse the post-processing of Non-Maximum-Suppression (NMS).}
  \label{fig:1}
  \vskip -0.10in
\end{figure}

Supervised learning has advanced object detection in the past few years, benefited from tremendous labeled training data \cite{ren2017faster,redmon2017yolo9000,law2020cornernet,tian2019fcos,chen2019all}. However, it is extremely expensive and time-consuming to collect accurate annotations. As an alternative, semi-supervised object detection (SSOD) is proposed to use a small amount of labeled data in conjunction with a large amount of unlabeled data to optimize the detectors \cite{jeong2019consistency, zhou2021instant, liu2021unbiased, sohn2020a, xu2021end}. Recently, SSOD has achieved growing interest in the object detection community.

Self-training has been proven useful in SSOD, especially mean teacher framework \cite{liu2021unbiased,xu2021end}, which annotates the unlabeled data by a gradually evolving teacher and guides the learning of a student in a mutually beneficial manner. As the key process of mean teacher, the existing pseudo labeling methods \cite{liu2021unbiased, zhou2021instant, xu2021end} simply utilize a hand-crafted confidence threshold to filter out low-quality pseudo labels and directly treat the remaining ones as reliable pseudo labels. However, it is inevitable to encounter the label mismatch problem, leading to severe confirmation bias \cite{arazo2020pseudo} during self-training. In this paper, we delve into this problem from two perspectives, \emph{i.e.}, \emph{distribution-level} and \emph{instance-level}. 

From the perspective of the distribution-level label mismatch problem, it is extremely difficult to generate unbiased pseudo labels to match the ground-truth labels with consistent class distribution by using a single and fixed confidence threshold due to the class-imbalanced data distribution. As shown in \cref{fig:1}, the number of pseudo labels is much higher than the ground-truth labels in some classes, while far less in some other classes, resulting in abundant false positives and false negatives. From the perspective of the instance-level label mismatch problem, the existing methods directly follow supervised object detection \cite{ren2017faster} for label assignment. However, the situation is totally different in semi-supervised learning since the quality of pseudo label cannot be guaranteed, leading to label assignment ambiguity problem as illustrated in \cref{fig:1}. Especially in mean teacher driven self-training framework, it is crucial to study how to assign the pseudo labels generated by the mean teacher to the proposals generated by the student network rather than a rough IoU based label assignment manner \cite{ren2017faster}. Based on the aforementioned challenges of label mismatch in two different yet complementary granularities, we begin our study and develop a \emph{LabelMatch} framework.

To address the first challenge, we present a very simple \emph{re-distribution mean teacher}. Assumed that the labeled data is selected from the entire data gallery via Monte Carlo Sampling. In this way, the label distribution of the unlabeled data can be approximated from that of the labeled data. In fact, we have evaluated the label distribution of the labeled and unlabeled data in several popular SSOD datasets, and they all meet this hypothesis which can be exploited as a weakly supervision cue for pseudo labeling. Under this inspiration, in contrast to a single and fixed confidence threshold, we utilize \emph{adaptive label-distribution-aware confidence thresholds} (ACT) to generate unbiased pseudo labels for the unlabeled data, supervised by the label distribution in the labeled data. The ACT are category-specific and adaptively up-to-date during self-training.

To address the second challenge, we propose a novel \emph{proposal self-assignment} method. Before introducing our method, we should highlight that it is infeasible to set all pseudo labels as hard labels due to the poor quality of the pseudo labels, especially at the beginning of self-training. Under this consideration, we divide the pseudo labels into reliable ones and uncertain ones according to the confidence score. We treat the reliable labels as hard labels for model optimization identically to the supervised manner, while exploiting the uncertain ones via the proposal self-assignment method for soft learning. Detailedly, the proposals from the student are injected into the teacher for proposals correction, which can provide corresponding soft labels to rectify each proposal accordingly. Besides, to encourage positive feedback during self-training, we introduce a \emph{reliable pseudo label mining} (RPLM) strategy to further improve the performance, which aims to convert the high-quality uncertain pseudo labels into reliable ones in a curriculum way.

We benchmark LabelMatch with the same experimental settings to Unbiased-Teacher \cite{liu2021unbiased} using the MS-COCO \cite{lin2014microsoft} and PASCAL-VOC \cite{everingham2010the} datasets, namely \emph{COCO-standard}, \emph{COCO-additional}, and \emph{VOC}. LabelMatch achieves new state-of-the-art results across all benchmarks. Especially in the settings with scarce labeled data, \emph{i.e.}, \emph{COCO-standard} with only 1\% labeled data and \emph{VOC}, our method can surpass the previous state-of-the-arts by a large margin.

The contributions of this paper are listed as follows:
\begin{itemize}[leftmargin=12pt, topsep=2pt, itemsep=0pt]
\item We contribute to analyzing the label mismatch problem from the perspectives of distribution-level and instance-level, which provides a brand-new direction for SSOD.
\item We propose a simple yet effective LabelMatch framework to address the label mismatch problems in SSOD. In this framework, we 1) present a re-distribution mean teacher to address the distribution-level label mismatch problem; 2) design a proposal self-assignment scheme to address the instance-level label mismatch problem; 3) introduce a reliable pseudo label mining strategy for pseudo label re-calibration during self-training.
\item The LabelMatch framework achieves new state-of-the-arts on many popular SSOD benchmarks. Also, we build a MMDetection-based semi-supervised object detection codebase for the fair study of SSOD algorithms.
\end{itemize}

\begin{figure*}
  \centering
    \includegraphics[width=\linewidth]{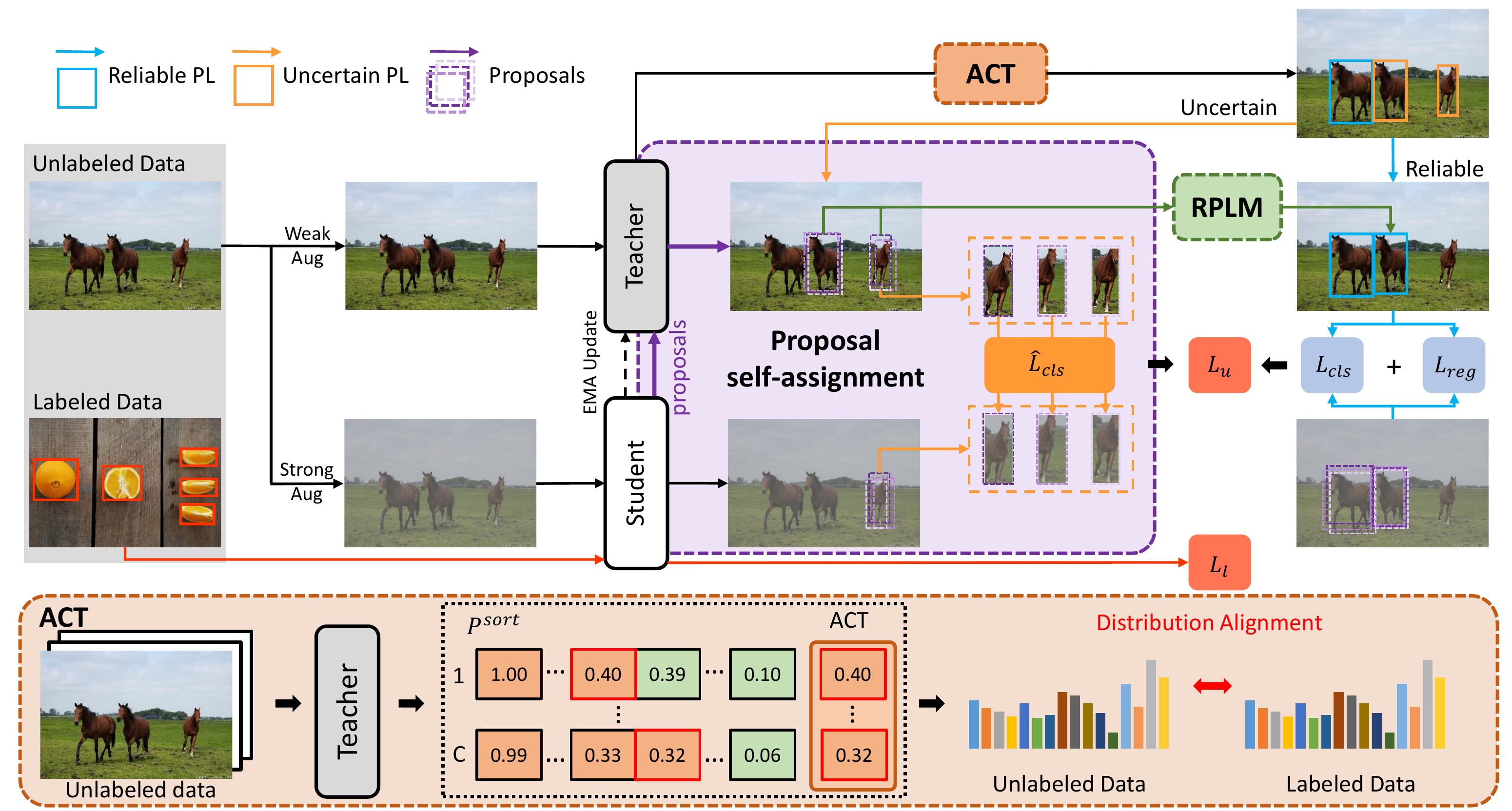}
  \vskip -0.1in
  \caption{An overview of LabelMatch framework. Labeled data: only applied to the student with a supervised loss. Unlabeled data: annotated by the teacher to get pseudo labels (PL) according to the \emph{adaptive label-distribution-aware confidence thresholds} (ACT), which are then split into reliable ones and uncertain ones for separated optimization. Reliable pseudo labels directly follow the IoU based assignment strategy, acting as hard labels to train the student model. As for uncertain labels, the \emph{proposal self-assignment} method guides the student training with the supervision provided by the corresponding proposal prediction in the teacher. Besides, a \emph{reliable pseudo label mining} (RPLM) strategy is utilized to convert the high-quality uncertain pseudo labels into reliable ones as the training goes on.}
  \label{fig:2}
  \vskip -0.1in
\end{figure*}

\section{Related Work}
\label{sec:related work}

\noindent
{\bf Semi-Supervised Classification.} The general methods can be roughly categorized into two types. One is consistency regularization, assuming the model's predictions to be invariant even if various perturbations are applied. There are different kinds of perturbations, including model-level perturbations \cite{sajjadi2016regularization, tarvainen2017mean, izmailov2018averaging}, image augmentations \cite{xie2020unsupervised}, and adversarial training \cite{miyato2019virtual}. Another one is self-training, aka pseudo labeling, which regards the predictions as pseudo labels. For instance, NoisyStudent \cite{xie2020self} evolves the pseudo labels for model optimization iteratively. MixMatch \cite{berthelot2019mixmatch} uses mixup augmentation and averages different augmented predictions to generate pseudo labels. FixMatch \cite{sohn2020fixmatch} uses the weakly augmented data for pseudo labeling while exploiting the strongly augmented data for model training.

\vspace{3mm}
\noindent{\bf Semi-Supervised Object Detection.} The technologies in SSOD are inherited from semi-supervised classification, dividing into consistency regularization \cite{jeong2019consistency, tang2021proposal} and self-training \cite{sohn2020a, liu2021unbiased, zhou2021instant, yang2021interactive, xu2021end}. In this paper, we mainly focus on the latter one. STAC \cite{sohn2020a} first generates pseudo labels by the pre-trained model and then feeds them back into the network with strong augmentation for model fine-tuning. To simplify this offline pseudo labeling, mean teacher based methods \cite{zhou2021instant, liu2021unbiased, xu2021end} perform a weak data transformation for online pseudo labeling and a strong data transformation for model training. However, there lies both foreground-background imbalance and foreground classes imbalance in SSOD, which makes it a more challenging task than semi-supervised classification. Unbiased-Teacher \cite{liu2021unbiased} and Soft-Teacher \cite{xu2021end} use the focal-loss and the soft-weight to alleviate these problems, respectively. Despite the great progress, the label mismatch problem during pseudo labeling still exists in the previous works. In contrast, we propose a LabelMatch framework to solve this problem from perspectives of distribution-level and instance-level.

\noindent{\bf Label Assignment.} It is necessary to assign the target of classification and localization for each proposal or anchor in object detection, known as label assignment \cite{ge2021ota}, which can be categorized into fixed and dynamic variants \cite{ge2021ota}. IoU based and center based label assignment are two common fixed assigning strategies, the first of which shows effectiveness in both RCNN-series \cite{girshick2014rich, girshick2015fast, ren2017faster, dai2016r, pang2019libra} and one-stage detectors \cite{redmon2017yolo9000, redmon2018yolov3, lin2020focal, liu2016ssd}, while the second one is popular in many anchor-free object detection \cite{redmon2016you, tian2019fcos, kong2020foveabox}. Recently, many adaptive mechanisms have been proposed to promote the label assignment, such as ATSS \cite{zhang2020bridging}, PAA \cite{kim2020probabilistic}, AutoAssign \cite{zhu2020autoassign}, OTA \cite{ge2021ota}, etc. However, all of these methods are only applied in supervised object detection, leaving a blank in SSOD due to the complex situation. To cope with the instance-level label mismatch problem, a novel label assignment is proposed to facilitate self-training in this paper.

\section{Methodology}
\label{sec:methodology}

In SSOD, a set of labeled images ${D_l}$$=$$\{x_i^l, y_i^l\}_{i=1}^{N_l}$ and a set of unlabeled images ${D_u}$$=$$\{x_i^u\}_{i=1}^{N_u}$ are provided, where $N_l$ and $N_u$ represent the number of labeled and unlabeled data, respectively. The annotation $y_i^l$ contains both categories and bounding boxes information. 

\subsection{Overview}
\label{subsec: overview}

The pipeline of LabelMatch framework is illustrated in \cref{fig:2}, which is derived from a basic mean teacher framework. The main idea of the mean teacher framework is to drive the teacher and student to evolve in a mutual learning mechanism. However, previous mean teacher based works inevitably suffer from the label mismatch problems, which we divide into two granularities, including distribution-level and instance-level. To solve these problems, we modify the mean teacher framework and develop a LabelMatch framework, consisting of a re-distribution mean teacher to solve the distribution-level label mismatch problem and a proposal self-assignment method to deal with the instance-level label mismatch problem. Also, it is beneficial to explore more high-quality pseudo labels. Thus, we further equip the proposed LabelMatch with a reliable pseudo label mining strategy to improve performance. 

\subsection{Preliminary: Mean Teacher Framework}
\label{subsec: basic}

Our approach follows the regimen of mean teacher, which contains a teacher model for pseudo label generation and a student model to improve the teacher model by updating knowledge. Both labeled and unlabeled data jointly constitute the batch of data. In each iteration, the teacher model first generates pseudo labels on the weakly-augmented unlabeled data, which are served as supervision signals for the corresponding strongly-augmented version. Subsequently, the student model is trained on the labeled data and the strongly-augmented unlabeled data with pseudo labels. In this way, the final training objective consists of a supervised loss and an unsupervised loss:
\begin{small}
\begin{equation}
\begin{aligned}
\mathcal{L}_l=\sum_i \mathcal{L}_{cls}(x_i^l, y_i^l)+\mathcal{L}_{reg}(x_i^l, y_i^l),
 \label{eq:1}
\end{aligned}
\end{equation}
\end{small}\begin{small}
\begin{equation}
\begin{aligned}
\mathcal{L}_u=\sum_i \mathcal{L}_{cls}(x_i^u, y_i^u)+\mathcal{L}_{reg}(x_i^u, y_i^u),
 \label{eq:2}
\end{aligned}
\end{equation}
\end{small}where $\mathcal{L}_{cls}$ is the classification loss, $\mathcal{L}_{reg}$ is the box regression loss, and $y_i^u$ is the pseudo label annotated by the teacher model. The overall loss is defined as:
\begin{small}
\begin{equation}
\mathcal{L}_{total}=\mathcal{L}_l+\lambda \mathcal{L}_u,
  \label{eq:3}
\end{equation}
\end{small}where $\lambda$ is a weight to balance the unsupervised loss, which is set 2.0 by default in this paper. During self-training, the teacher gradually updates its weights from the student via an exponential moving average (EMA) strategy.

\subsection{LabelMatch}
\label{subsec: labelmatch}

We claim that the main obstacle to hinder the performance of mean teacher framework lies in the label mismatch problem. The proposed LabelMatch framework adopts the same mean teacher scheme, but develops a re-distribution mean teacher, which utilizes adaptive label distribution-aware confidence thresholds (ACT) to achieve unbiased pseudo labels. Moreover, a proposal self-assignment method and a reliable pseudo label mining strategy are introduced to rectify self-training.

\vspace{2mm}
\noindent
{\bf Re-distribution Mean Teacher.} In semi-supervised learning, the labeled data and the unlabeled data are from the same data distribution. Intuitively, we can obtain adaptive thresholds by minimizing the discrepancy of class distributions between the labeled and the unlabeled data, which can be formulated as follows:
\begin{small}
\begin{equation}
\begin{aligned}
\underset{t_1, ..., t_C}{\mathrm{argmin}} \quad & D_{KL}([\underbrace{r_1^l, ..., r_C^l}_{f-f}, \underbrace{r_{f}^l}_{f-b}], [\underbrace{r_1^u, ..., r_C^u}_{f-f}, \underbrace{r_f^u}_{f-b}]) \\
\textrm{s.t.} \quad
& r_c^l=\frac{n_c^l}{\sum_{i=1}^C n_i^l}, \\
& r_f^l=\frac{\sum_{i=1}^C n_i^l}{N_l}, \\
& r_c^u=\frac{\sum_{j=1}^{N_u}\sum(P_c^j>t_c)}{\sum_{i=1}^C(\sum_{j=1}^{N_u}\sum (P_i^j>t_i))}, \\
& r_f^u=\frac{\sum_{i=1}^C(\sum_{j=1}^{N_u}\sum (P_i^j>t_i))}{N_u}, \\
\end{aligned}
\label{eq:4}
\end{equation}
\end{small}where $D_{KL}$ represents the Kullback-Leibler divergence between two distributions, $n_i^l$ denotes the box number of the $i$-th class in the labeled data, $C$ is the entire foreground class number, $P_i^j$ is a list of prediction scores of the $i$-th class in the $j$-th unlabeled image, $f$-$f$ means the foreground-foreground class distribution, and $f$-$b$ means the foreground-background ratio. Note that all the predictions in the unlabeled data are post-processed by NMS. $t_c$ is the optimized variant, aka the confidence threshold to filter pseudo boxes for the $c$-th category, determined as:
\begin{small}
\begin{equation}
t_c=P_c^{sort}[n_c^l\cdot \frac{N_u}{N_l}],
\label{eq:5}
\end{equation}
\end{small}where $P_c^{sort}$ is a list of prediction scores of the $c$-th class, which are sorted by descending. For efficient implementation, only a subset of unlabeled data are selected to estimate the distribution for thresholds determination. While the model is consecutively optimized during training, the previous thresholds are however imprecise for pseudo labeling, failing to be consistent with the truth class distribution. We thus simply update these thresholds every $K$ iterations to dynamically adjust to the current teacher model. Such that, the thresholds are category-specific and adaptively up-to-date, termed as adaptive label-distribution-aware confidence thresholds (ACT), which we identify as the critical step to solve the distribution-level mismatch problem.

\vspace{2mm}
\noindent
{\bf Proposal Self-Assignment.} It is worth noting that the quality of pseudo labels cannot be guaranteed, especially at the early beginning of self-training. Inspired by the noise label learning \cite{han2018co,SSNLL}, we divide pseudo labels into reliable ones and uncertain ones according to the confidence score. Denoting $\alpha\%$ as the pre-defined proportion of reliable pseudo labels, the confidence thresholds $t_c^r$ to filter reliable pseudo labels for the $c$-th category can be written as:
\begin{small}
\begin{equation}
t_c^r=P_c^{sort}[\alpha\%\cdot n_c^l\cdot \frac{N_u}{N_l}],
 \label{eq:6}
\end{equation}
\end{small}Pseudo labels with confidence higher than $t_c^r$ are regarded as hard labels for student model optimization in a supervised manner. In contrast, the remaining uncertain ones are treated as soft labels for soft learning.

Obviously, the uncertain pseudo labels potentially lead to low localization accuracy. To avoid poor box regression, Unbiased-Teacher removes the box regression loss for the unlabeled data, yet resulting in ambiguity in label assignment as shown in \cref{fig:1}. For example, supposed that the proposals with an IoU overlap higher than 0.5 are optimized to the same uncertain pseudo labels, they will tend to be the same in classification score but different in localization after being refined by ROIHead. These refined proposals behave indistinguishably for the NMS post-processing, which confuses NMS to suppress redundant boxes randomly. We refer to this situation as an instance-level label mismatch problem, lacking attention in the previous SSOD works. To this end, we present a novel proposal self-assignment method for proposal re-calibration. Specifically, we utilize the proposals matched to the uncertain pseudo labels generated by the student to extract the corresponding features in the teacher, and then feed these features into the ROIHead of the teacher to achieve the refined boxes. Different from the IoU based label assignment, each proposal in the student uses the corresponding soft labels refined by the ROIHead of the teacher model for self-training, and finally, varying from each other in classification score to avoid NMS confusion. In this way, we optimize the student model with the uncertain pseudo labels via a soft classification loss:
\begin{small}
\begin{equation}
\hat{\mathcal{L}}_{cls}=\sum_{i=1}^{n_p}\sum_{c=1}^C-p^t_{i,c}\log p^s_{i,c},
  \label{eq:7}
\end{equation}
\end{small}where $n_p$ is the number of the corresponding proposals matched to the uncertain pseudo labels, $C$ is the class number, $p_{i,c}^s$ is the probability of the $c$-th class in the $i$-th proposal from the student model, and $p_{i,c}^t$ denotes the corresponding soft label from the teacher model matched to $p_{i,c}^s$.

Incorporating the re-distribution mean teacher and the proposal self-assignment for pseudo labeling, the unsupervised loss in \cref{eq:2} can be reformulated as:
\begin{small}
\begin{equation}
\mathcal{L}_u\!\!=\!\!\sum_i \mathcal{L}_{cls}(x_i^u,y_i^{ur})\!\!+\!\!\mathcal{L}_{reg}(x_i^u,y_i^{ur})\!\!+\!\!\hat{\mathcal{L}}_{cls}(x_i^u,y_i^{uu}),
\label{eq:8}
\end{equation}
\end{small}where $y_i^{ur}$ and $y_i^{uu}$ denote the reliable pseudo labels and the uncertain pseudo labels, respectively.

\begin{figure}[t]
  \centering
  \begin{subfigure}{0.495\linewidth}
    \includegraphics[width=\linewidth]{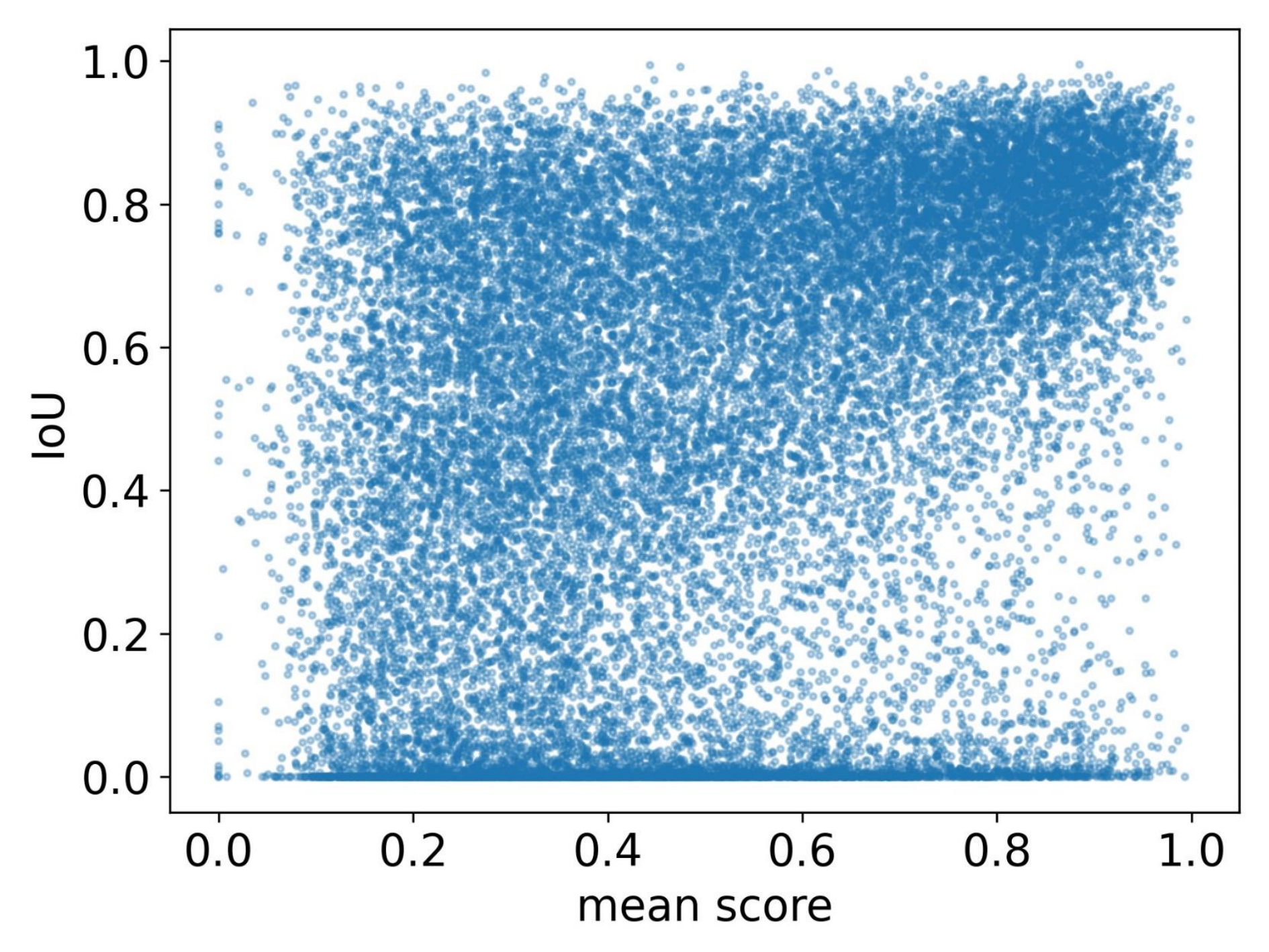}
  \vskip -0.05in
    \caption{}
    \label{fig:3-a}
  \end{subfigure}
  \hfill
  \begin{subfigure}{0.495\linewidth}
    \includegraphics[width=\linewidth]{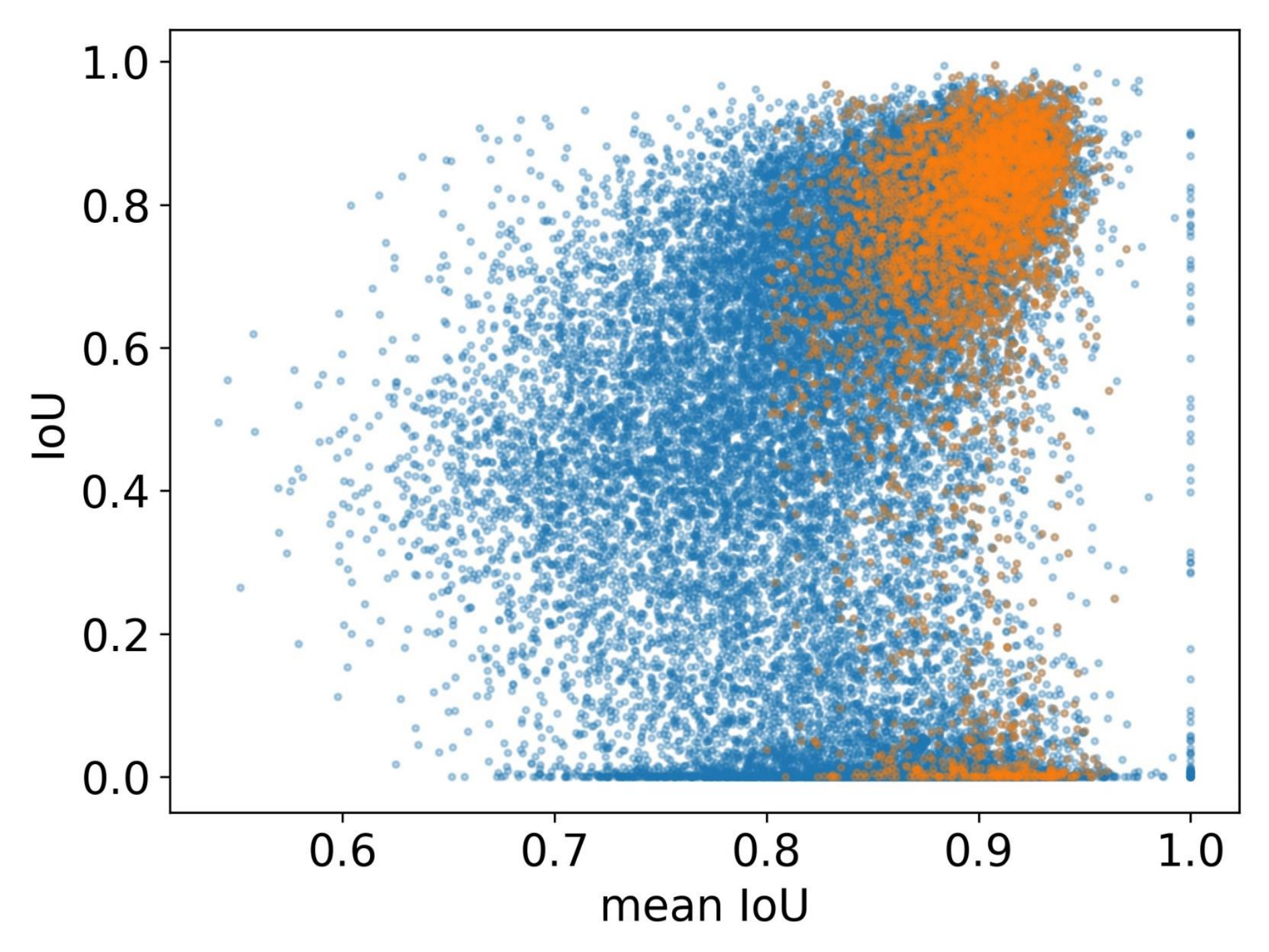}
  \vskip -0.05in
    \caption{}
    \label{fig:3-b}
  \end{subfigure}
  \vskip -0.1in
  \caption{5k images are selected to estimate the quality of pseudo labels. (a) the correlation between the IoU with ground truth and \emph{mean score}. (b) the correlation between the IoU with ground truth and \emph{mean IoU}. The orange points represent the predictions with the \emph{mean score} larger than 0.8 and \emph{mean IoU} larger than 0.8}
  \vskip -0.15in
  \label{fig:3}
\end{figure}

\vspace{2mm}
\noindent
{\bf Reliable Pseudo Label Mining.} To benefit from the continuously evolved teacher model and encourage the cycle positive feedback during self-training, we present a reliable pseudo label mining strategy to convert the high-quality uncertain pseudo labels into reliable ones. First, it is known that a set of adjacent boxes will be suppressed into one box after NMS. In this paper, we claim that these adjacent boxes before NMS can be exploited to evaluate the quality of the corresponding pseudo label after NMS. In this way, we present two evaluation metrics in this paper, \emph{i.e.}, \emph{mean score} and \emph{mean IoU}, which are the average classification scores and IoU of this set of adjacent boxes matched to the corresponding bounding box after NMS. Note that We use the predictions from the teacher to compute \emph{mean score} and \emph{mean IoU}, and the IoU scores are determined by the IoU between the suppressed boxes and the selected box in NMS. We claim that the pseudo labels with higher quality usually correspond to higher mean scores and higher mean IoU. The empirical study in \cref{fig:3} gives a demonstration of our hypothesis. In this way, the uncertain pseudo labels with mean scores larger than $T_{score}$ and mean IoU larger than $T_{iou}$ will be transferred to reliable pseudo labels.

\begin{table*}
  \centering
  \resizebox{1.0\textwidth}{!}{
  \begin{tabular}{p{0.2\textwidth}>{\centering}p{0.1\textwidth}>{}p{0.2\textwidth}>{}p{0.2\textwidth}>{\arraybackslash}p{0.2\textwidth}}
    \toprule
    & Threshold & \multicolumn{1}{c}{1\%} & \multicolumn{1}{c}{5\%} & \multicolumn{1}{c}{10\%} \\
    \midrule
    Supervised \cite{liu2021unbiased} & - & $9.05\pm 0.16$ & $18.47\pm 0.22$ & $23.86\pm 0.81$ \\
    \hline
    STAC \cite{sohn2020a} & 0.9 & $13.97\pm 0.35 $ \small{(\textcolor{blue}{$+4.92$})} & $24.38\pm 0.12 $ \small{(\textcolor{blue}{$+5.91$})} & $28.64\pm 0.21 $ \small{(\textcolor{blue}{$+4.78$})} \\
    ISMT \cite{yang2021interactive} & 0.9 & $18.88\pm 0.74 $ \small{(\textcolor{blue}{$+9.83$})} & $26.37\pm 0.24 $ \small{(\textcolor{blue}{$+7.90$})} & $30.53\pm 0.52 $ \small{(\textcolor{blue}{$+6.67$})} \\
    Instant Teaching\cite{zhou2021instant} & 0.9 & $18.05\pm 0.15 $ \small{(\textcolor{blue}{$+9.00$})} & $26.75\pm 0.05 $ \small{(\textcolor{blue}{$+8.28$})} & $30.40\pm 0.05 $ \small{(\textcolor{blue}{$+6.54$})} \\
    Unbiased Teacher \cite{liu2021unbiased} & 0.7 & $20.75\pm 0.12 $ \small{(\textcolor{blue}{$+11.70$})} & $28.27\pm 0.11 $ \small{(\textcolor{blue}{$+9.80$})} & $31.50\pm 0.10 $ \small{(\textcolor{blue}{$+7.64$})} \\
    Soft Teacher \cite{xu2021end} & 0.9 & $20.46\pm 0.39 $ \small{(\textcolor{blue}{$+11.41$})} & $30.74\pm 0.08 $ \small{(\textcolor{blue}{$+12.27$})} & $34.04\pm 0.14 $ \small{(\textcolor{blue}{$+10.18$})} \\
    LabelMatch (Ours) & ACT & ${\bf 25.81\pm 0.28}$ \small{(\textcolor{blue}{$+16.76$})} & ${\bf 32.70\pm 0.18}$ \small{(\textcolor{blue}{$+14.23$})} & ${\bf 35.49\pm 0.17}$ \small{(\textcolor{blue}{$+11.63$})} \\
    \bottomrule
  \end{tabular}
  }
    \vskip -0.1in
  \caption{Experimental results on COCO-standard ($AP_{50:95}$). All the results are the average of all 5 folds.}
  \label{tab:1}
    \vskip -0.1in
\end{table*}

\section{Experiments}
\label{set: exp}

\subsection{Experimental Setup}
\noindent
{\bf Datasets.} We evaluate our method on the MS-COCO \cite{lin2014microsoft} and PASCAL-VOC \cite{everingham2010the} datasets. There are three settings following the existing works \cite{liu2021unbiased,sohn2020a}: (1) COCO-standard: 1\%, 5\%, 10\% images of \texttt{train2017} set are sampled as the labeled training data and the remaining ones as the unlabeled data. (2) COCO-additional: we use the entire \texttt{train2017} set as the labeled data and the additional \texttt{COCO2017-unlabeled} set as the unlabeled data. (3) VOC: we use VOC07 \texttt{trainval} set as the labeled data and the VOC12 \texttt{trainval} set as the unlabeled data. The validation sets in the COCO setting and VOC setting are COCO \texttt{val2017} and VOC07 test set, respectively.

\vspace{1mm}
\noindent
{\bf Network.} For a fair comparison, we use Faster-RCNN \cite{ren2017faster} with FPN \cite{lin2017feature} and ResNet-50 backbone \cite{he2016deep} as the detector. Our framework can be easily extended to other detectors.

\vspace{1mm}
\noindent
{\bf Implementation Details.} We implement our method with MMDetection \cite{chen2019mmdetection}. For data augmentation, we apply random horizontal flipping and multi-scale for weak augmentation. Based on this augmentation, we then add random color jittering, grayscale, gaussian blurring and cutout patches for strong augmentation, which is similar to \cite{liu2021unbiased}. The $T_{score}$ and $T_{iou}$ in RPLM are set to 0.8 by default. More training and implementation details are introduced in the Appendix. 

\begin{table}
  \centering
  \resizebox{0.48\textwidth}{!}{
  \begin{tabular}{p{0.19\textwidth}>{\centering}p{0.07\textwidth}>{\centering\arraybackslash}p{0.14\textwidth}}
    \toprule
    & Iterations & $AP_{50:95}$ \\
    \midrule
    STAC \cite{sohn2020a} & 540k & $39.5\xrightarrow{{-0.3}} 39.2 $ \\
    Unbiased Teacher \cite{liu2021unbiased} & 270k & $40.2\xrightarrow{{+1.1}} 41.3 $ \\
    Soft Teacher \cite{xu2021end} & 370k & $40.9\xrightarrow{{+3.6}} 44.5 $ \\
    LabelMatch (Ours) & 540k & $40.3\xrightarrow{{+5.0}} {\bf 45.3} $ \\
    \bottomrule
  \end{tabular}}
  \vskip -0.1in
  \caption{Experimental results on COCO-additional.}
  \label{tab:3}
\end{table}

\begin{table}
  \centering
  \resizebox{0.48\textwidth}{!}{
    \begin{tabular}{@{}lll@{}}
    \toprule
     & \multicolumn{1}{c}{$AP_{50}$} & \multicolumn{1}{c}{$AP_{50:95}$} \\
    \midrule
    Supervised \cite{liu2021unbiased} & 72.63 & 42.13 \\
    \hline
    STAC \cite{sohn2020a} & 77.45 \small{(\textcolor{blue}{$+4.82$})} & 44.64 \small{(\textcolor{blue}{$+2.51$})} \\
    ISMT \cite{yang2021interactive} & 77.23 \small{(\textcolor{blue}{$+4.60$})} & 46.23 \small{(\textcolor{blue}{$+4.10$})} \\
    Instant Teaching\cite{zhou2021instant} & 79.20 \small{(\textcolor{blue}{$+6.57$})} & 50.00 \small{(\textcolor{blue}{$+7.87$})} \\
    Unbiased Teacher \cite{liu2021unbiased} & 77.37 \small{(\textcolor{blue}{$+4.74$})} & 48.69 \small{(\textcolor{blue}{$+6.56$})} \\
    LabelMatch (Ours) & {\bf 85.48} \small{(\textcolor{blue}{$+12.85$})} & {\bf 55.11} \small{(\textcolor{blue}{$+12.98$})} \\
    \bottomrule
  \end{tabular}}
    \vskip -0.1in
  \caption{Experimental results on VOC.}
  \label{tab:2}
    \vskip -0.1in
\end{table}

\subsection{Results}


\vspace{1mm}
\noindent
{\bf COCO-standard.} We evaluate the proposed method on COCO-standard (\cref{tab:1}). Our method consistently outperforms the previous state-of-the-arts with different percentages of labeled data. It is worth mentioning that the proposed method achieves 25.81 mAP on 1\% labeled data, which is even higher than the supervised baseline trained on 10\% labeled data.

\vspace{1mm}
\noindent
{\bf COCO-additional.} We verify whether the model trained on 100\% COCO data can be further improved by using additional unlabeled COCO data. As shown in \cref{tab:3}, our method boosts the supervised baseline with +5.0 mAP,  while the existing SOTA improvement is +3.6 mAP.

\vspace{1mm}
\noindent
{\bf VOC.} We evaluate the proposed method on PASCAL-VOC. As is shown in \cref{tab:2}, our method achieves 85.48 mAP on $AP_{50}$, which outperforms previous state-of-the-arts by +6.28 absolute mAP improvement.

\subsection{Ablation Studies}
\label{ablation}

In ablation studies, we conduct experiments in the setting of 1\% COCO-standard (one of 5 folds), without RPLM strategy if not specified. The training iterations are reduced to 40k for all of the experiments. More implementation details and ablation studies can be found in the Appendix.

\begin{figure*}
  \centering
  \begin{subfigure}{0.25\linewidth}
    \includegraphics[width=\linewidth]{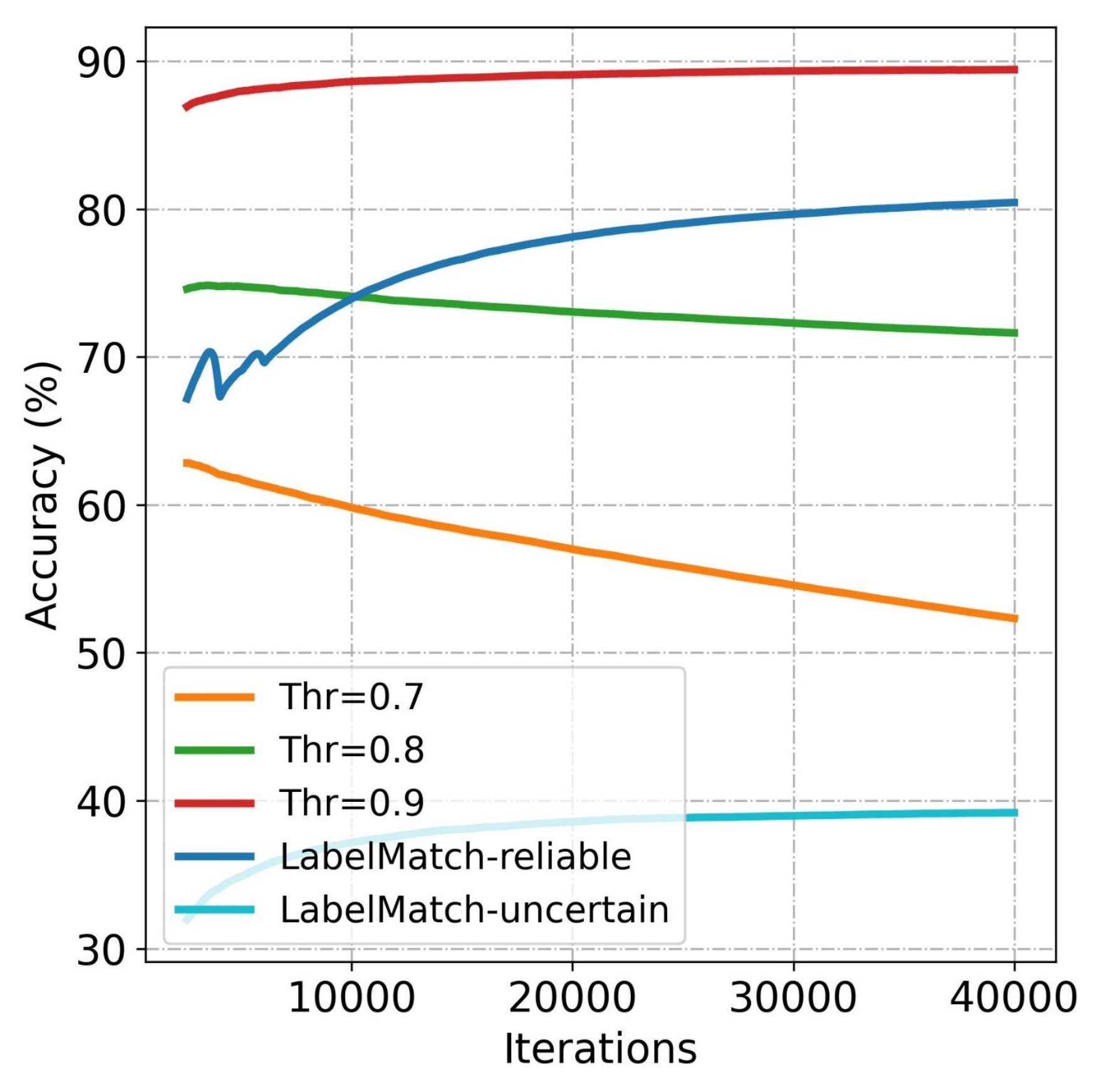}
    \caption{}
    \label{fig:4-a}
  \end{subfigure}
  \hfill
  \begin{subfigure}{0.25\linewidth}
    \includegraphics[width=\linewidth]{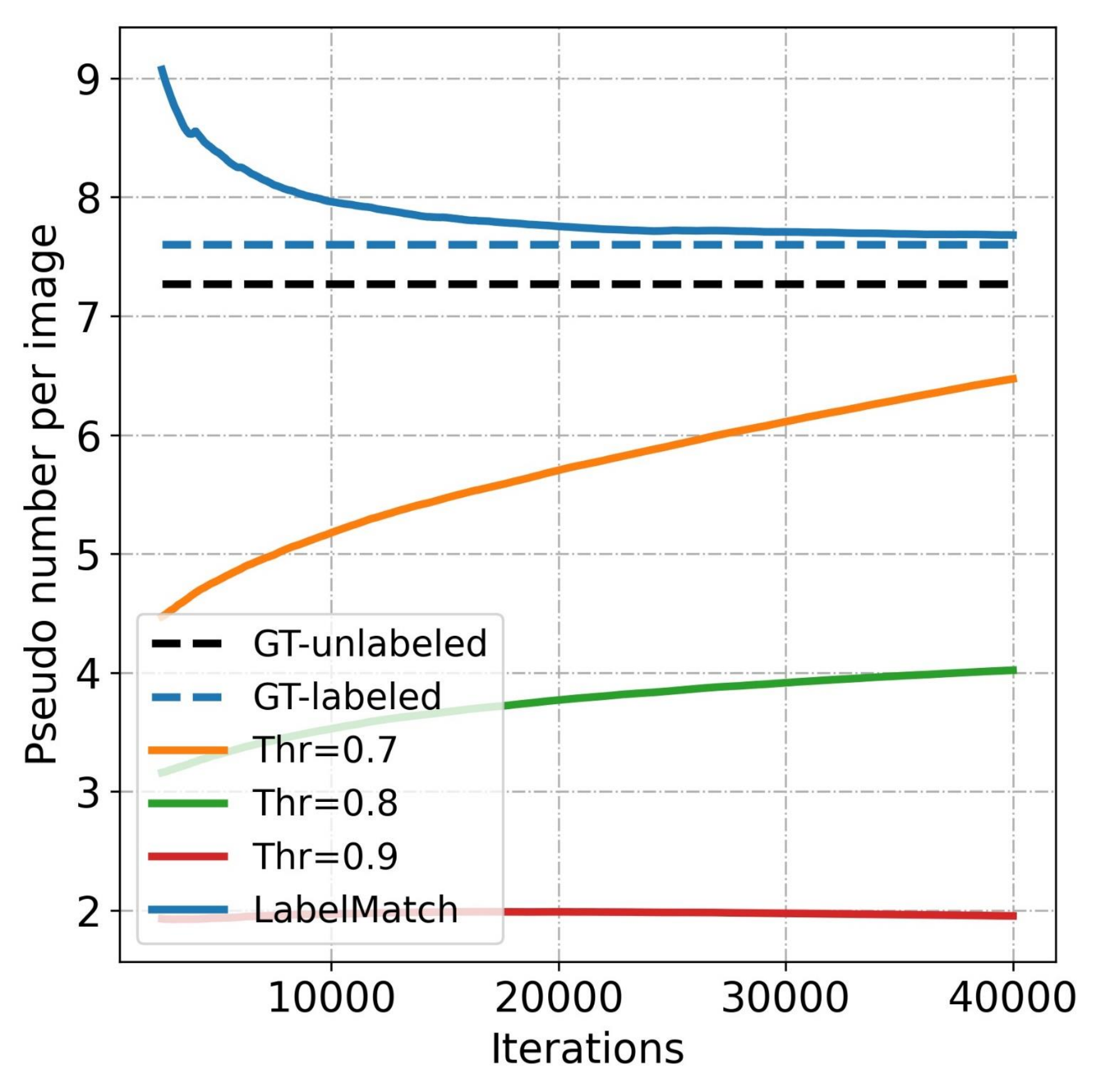}
    \caption{}
    \label{fig:4-b}
  \end{subfigure}
  \hfill
  \begin{subfigure}{0.49\linewidth}
      \includegraphics[width=\linewidth]{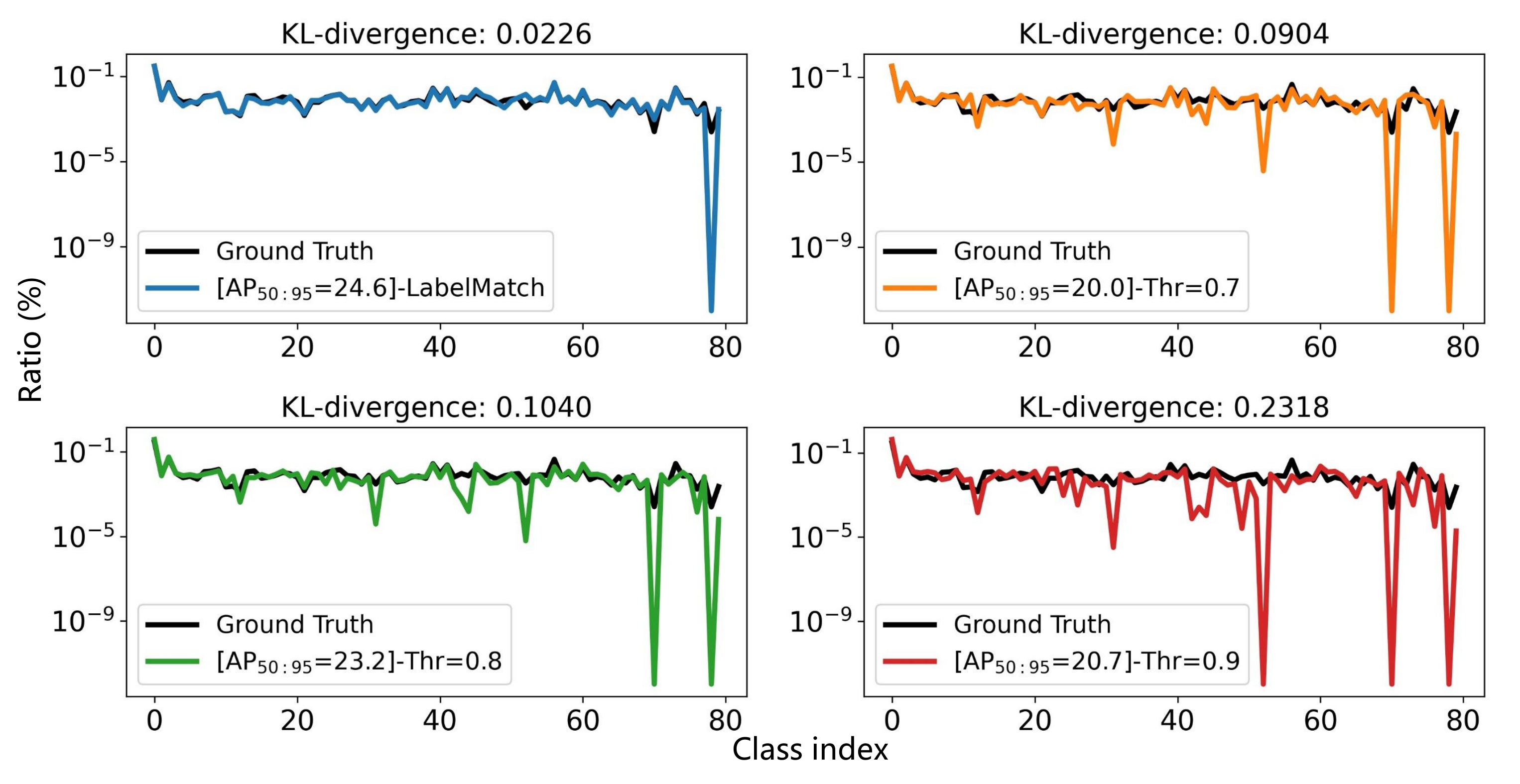}
    \caption{}
    \label{fig:4-c}
  \end{subfigure}
    \vskip -0.1in
  \caption{Ablation study about the quality of pseudo labels. (a) the accuracy of pseudo labels in the training phase. (Note: the pseudo labels with IoU overlapping the ground truth greater than 0.5 are regarded as true positives) (b) The average number of pseudo labels per image in the training phase. (c) KL divergence of the class distribution between the pseudo labels and the ground truth in the training phase.}
  \label{fig:4}
    \vskip -0.1in
\end{figure*}

\begin{table*}[!]
\resizebox{\textwidth}{!}{
\begin{tabular}{c|cccccccccccccccccccc|c}
\toprule
 & \rotatebox{90}{toaster} & \rotatebox{90}{hair drier} & \rotatebox{90}{scissors} & \rotatebox{90}{microwave} & \rotatebox{90}{toothbrush} & \rotatebox{90}{parking meter} & \rotatebox{90}{snowboard} & \rotatebox{90}{bear} & \rotatebox{90}{stop sign} & \rotatebox{90}{fire hydrant} & \rotatebox{90}{mous
 } & \rotatebox{90}{frisbee} & \rotatebox{90}{refrigerator} & \rotatebox{90}{hot dog} & \rotatebox{90}{oven} & \rotatebox{90}{airplane} & \rotatebox{90}{baseball bat} & \rotatebox{90}{baseball glove} & \rotatebox{90}{keyboard} & \rotatebox{90}{bed} & mean \\
 \midrule
thr=0.7 & 0.0 & \textbf{0.0} & 7.4 & 37.9 & \textbf{4.8} & 31.9 & 1.8 & 47.6 & 17.6 & 30.2 & 33.8 & 36.7 & 14.8 & 2.0 & 11.9 & 15.9 & 6.9 & 16.4 & 24.4 & 21.8 & 18.2 \\
thr=0.8 & 0.0 & 0.0 & 7.7 & 37.2 & 4.7 & 32.0 & 3.0 & \textbf{51.5} & 32.4 & 46.2 & 44.8 & 44.5 & 33.7 & 4.6 & \textbf{18.5} & 34.8 & 11.1 & \textbf{24.0} & 21.8 & 27.1 & 24.0 \\
thr=0.9 & 0.0 & 0.0 & 5.3 & \textbf{37.9} & 2.2 & 17.6 & 2.8 & 48.4 & \textbf{53.4} & 46.8 & 44.8 & 37.8 & 31.7 & 1.2 & 16.8 & 33.1 & 12.4 & 23.7 & 26.8 & 26.6 & 23.5 \\
\midrule
LabelMatch & \textbf{4.6} & 0.0 & \textbf{16.5} & 37.5 & 4.6 & \textbf{34.9} & \textbf{9.5} & 49.9 & 50.4 & \textbf{49.2} & \textbf{47.4} & \textbf{45.0} & \textbf{34.4} & \textbf{8.5} & 17.7 & \textbf{39.3} & \textbf{13.2} & 21.0 & \textbf{29.3} & \textbf{32.3} & \textbf{27.3} \\
\bottomrule
\toprule
& \rotatebox{90}{person} & \rotatebox{90}{car} & \rotatebox{90}{chair} & \rotatebox{90}{book} & \rotatebox{90}{bottle} & \rotatebox{90}{cup} & \rotatebox{90}{dining table} & \rotatebox{90}{traffic light} & \rotatebox{90}{bowl} & \rotatebox{90}{handbag} & \rotatebox{90}{bird} & \rotatebox{90}{boat} & \rotatebox{90}{truck} & \rotatebox{90}{umbrella} & \rotatebox{90}{bench} & \rotatebox{90}{cow} & \rotatebox{90}{banana} & \rotatebox{90}{carrot} & \rotatebox{90}{motorcycle} & \rotatebox{90}{backpack} & mean \\
\midrule
thr=0.7 & \textbf{39.1} & 30.5 & 9.2 & 2.1 & 20.3 & 24.0 & 12.5 & 19.4 & 22.2 & 4.3 & 16.1 & \textbf{14.4} & 3.5 & 20.0 & 7.7 & 31.8 & 8.9 & 3.1 & 27.4 & 3.8 & 16.0 \\
thr=0.8 & 39.0 & 31.8 & 10.3 & \textbf{2.1} & \textbf{25.8} & \textbf{26.3} & 14.1 & 20.2 & 25.6 & 4.1 & \textbf{18.1} & 13.6 & 8.6 & 20.0 & 11.4 & 32.1 & 8.3 & 3.9 & \textbf{28.0} & 4.8 & 17.4 \\
thr=0.9 & 32.0 & 26.6 & 7.1 & 1.0 & 13.6 & 19.9 & \textbf{14.9} & 17.8 & 23.0 & 1.1 & 14.9 & 8.1 & 12.7 & 15.4 & 10.3 & 24.2 & 6.6 & 1.8 & 24.5 & 2.5 & 13.9 \\
\midrule
LabelMatch & 38.8 & \textbf{31.9} & \textbf{11.7} & 1.9 & 25.6 & 26.1 & 14.6 & \textbf{20.8} & \textbf{29.2} & \textbf{4.8} & 17.9 & 13.2 & \textbf{14.8} & \textbf{21.2} & \textbf{13.6} & \textbf{39.0} & \textbf{9.1} & \textbf{4.9} & 27.4 & \textbf{6.1} & \textbf{18.6} \\
\bottomrule
\end{tabular}
}
  \vskip -0.1in
\caption{Quantitative results on top20 tail categories (upper) and top20 head categories (lower) ($AP_{50:95}$).}
\label{tab:4}
  \vskip -0.1in
\end{table*}

\begin{figure}
  \centering
  \includegraphics[width=\linewidth]{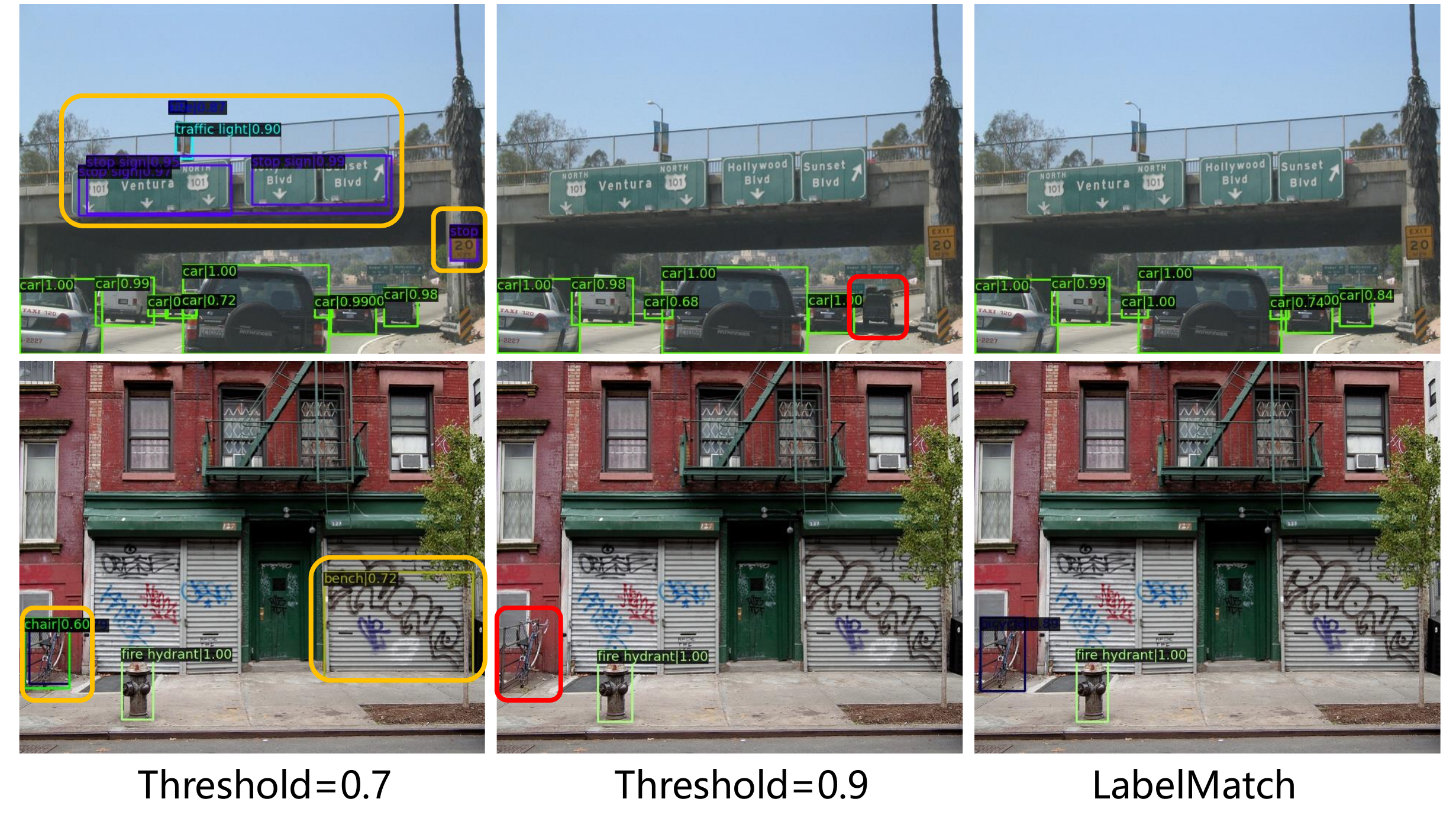}
    \vskip -0.1in
  \caption{Qualitative comparisons between the single confidence threshold and our proposed method. Red rectangles highlight the false negatives, and yellow rectangles highlight the false positives. The score threshold for visualization is 0.6.}
  \label{fig:5}
    \vskip -0.2in
\end{figure}

\vspace{3mm}
\noindent
{\bf The quality of pseudo labels.} The quality of pseudo labels can be reflected in three aspects: 1) the accuracy of pseudo labels; 2) foreground-background distribution; 3) class distribution (foreground-foreground distribution). We compare the proposed method against the single confidence threshold based mean teacher framework. LabelMatch shows superior advancement, which we attribute to the following aspects: 
\begin{itemize}[leftmargin=12pt, topsep=2pt, itemsep=0pt]
\item More accurate pseudo labels. As is shown in \cref{fig:4-a}, the accuracy of pseudo labels decreases when using threshold=0.7 and threshold=0.8. In contrast, the accuracy increases in LabelMatch. Although the accuracy achieves the best in threshold=0.9, the number (recall) of foregrounds is much less than the ground truth.
\item Unbiased foregrounds-background distribution. As is shown in \cref{fig:4-b}, the number of pseudo labels in our method is nearly the same as the ground truth, while the number in the single confidence threshold based method is much lower, especially when threshold=0.9. 
\item Consistent class distribution. \cref{fig:4-c} demonstrates that LabelMatch guarantees both the foreground-background distribution and the class distribution to be nearly consistent with the ground truth. The situation is totally different when using the single confidence threshold, which brings large gaps with ground truth in many categories.
\end{itemize}

To show the quality of pseudo labels more intuitively, we give the quantitative and qualitative demonstrations in \cref{tab:4} and \cref{fig:5}, respectively. From \cref{tab:4}, we observe obvious gains in AP on top20 tail and head categories by leveraging LabelMatch compared with the single and fixed threshold. As shown in \cref{fig:5}, there are many false positives with threshold=0.7 and many false negatives with threshold=0.9, which are removed by leveraging LabelMatch. These experimental results indicate the effectiveness of LabelMatch to re-distribute the pseudo label distribution, preventing self-training from collapsing to dominant classes. More qualitative results are shown in the Appendix.

\begin{figure*}
  \centering
  \begin{subfigure}{0.33\linewidth}
    \includegraphics[width=\linewidth]{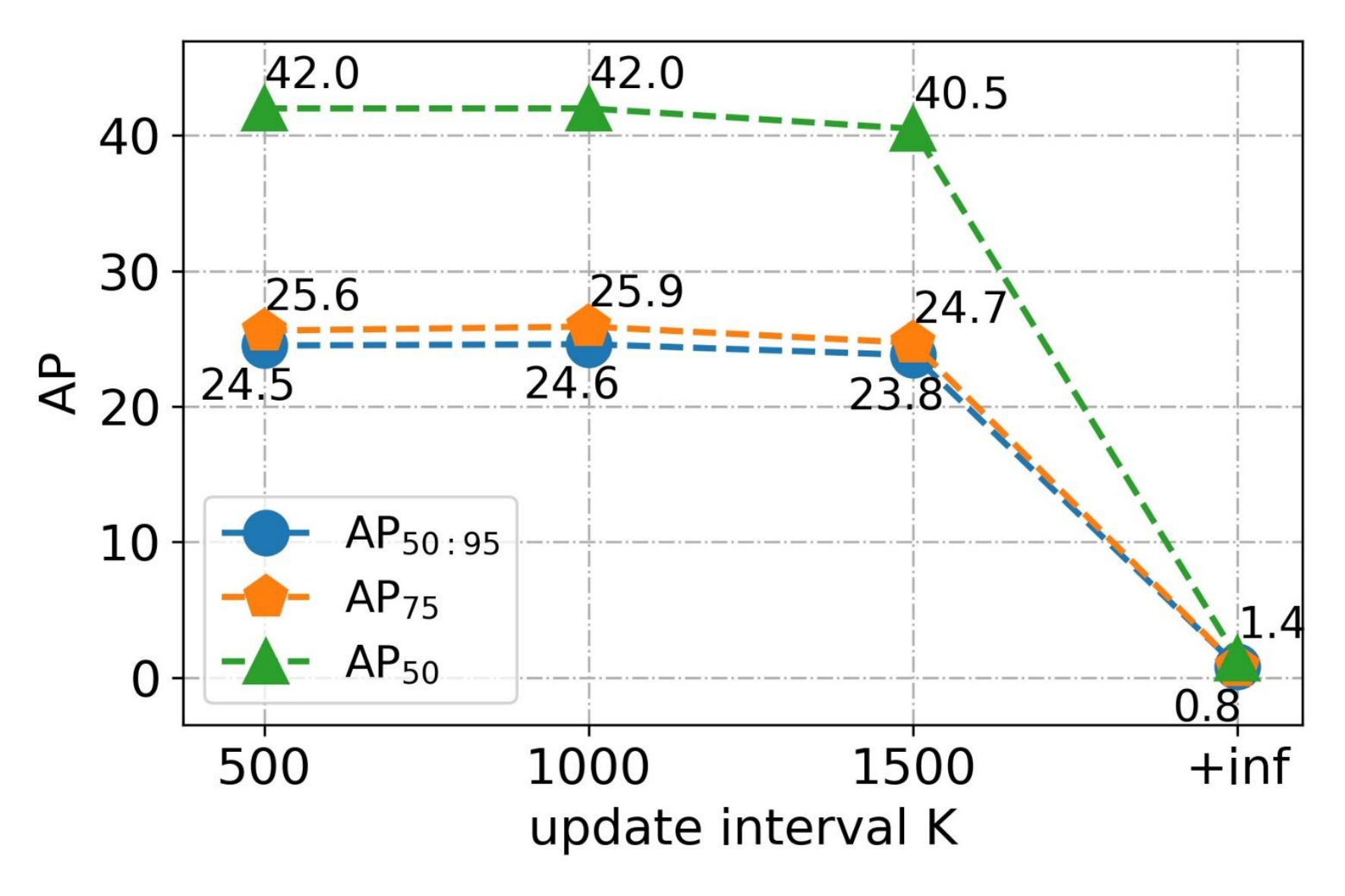}
    \caption{}
    \label{fig:6-a}
  \end{subfigure}
  \begin{subfigure}{0.33\linewidth}
    \includegraphics[width=\linewidth]{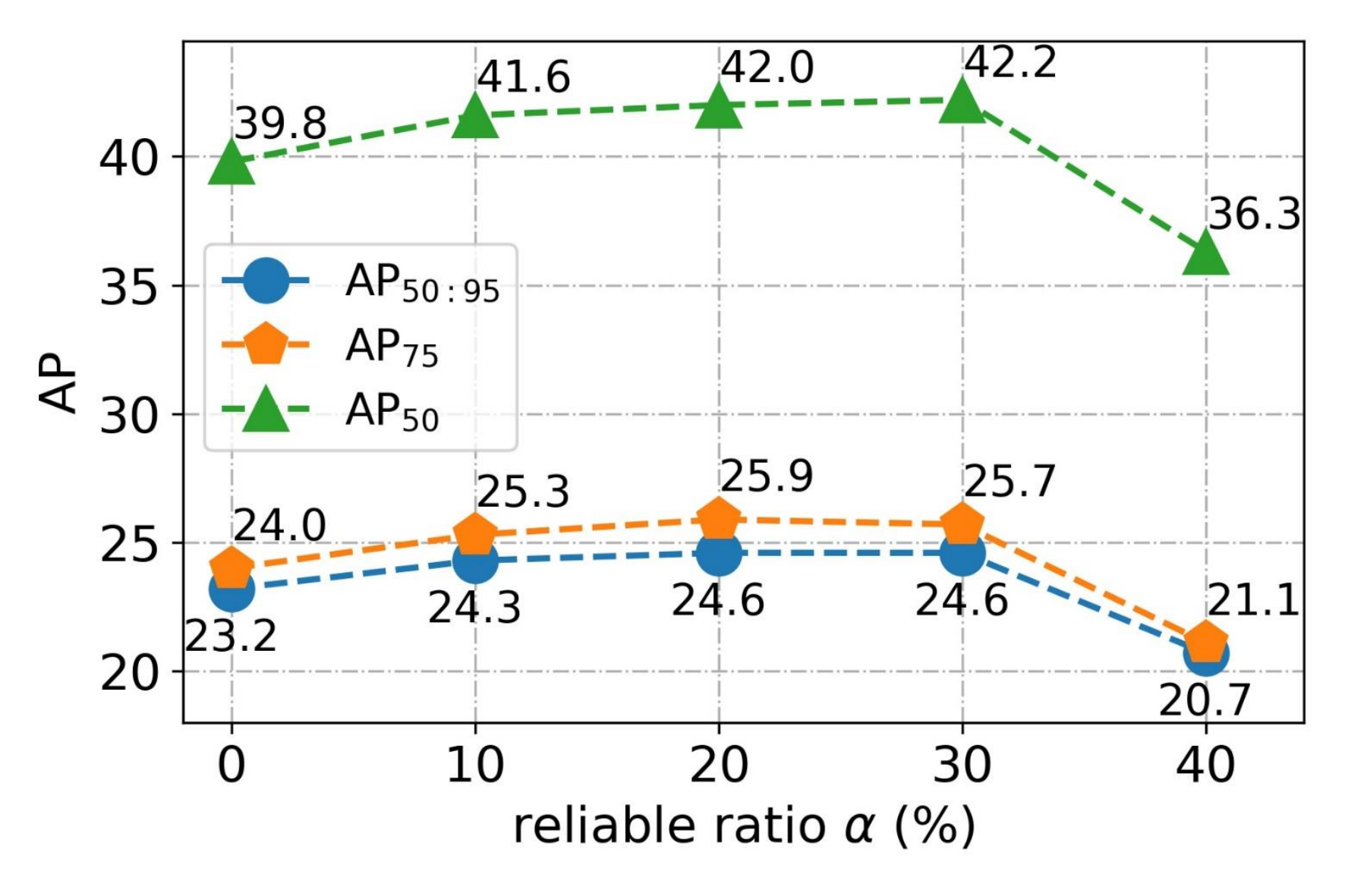}
    \caption{}
    \label{fig:6-b}
  \end{subfigure}
  \begin{subfigure}{0.33\linewidth}
    \includegraphics[width=\linewidth]{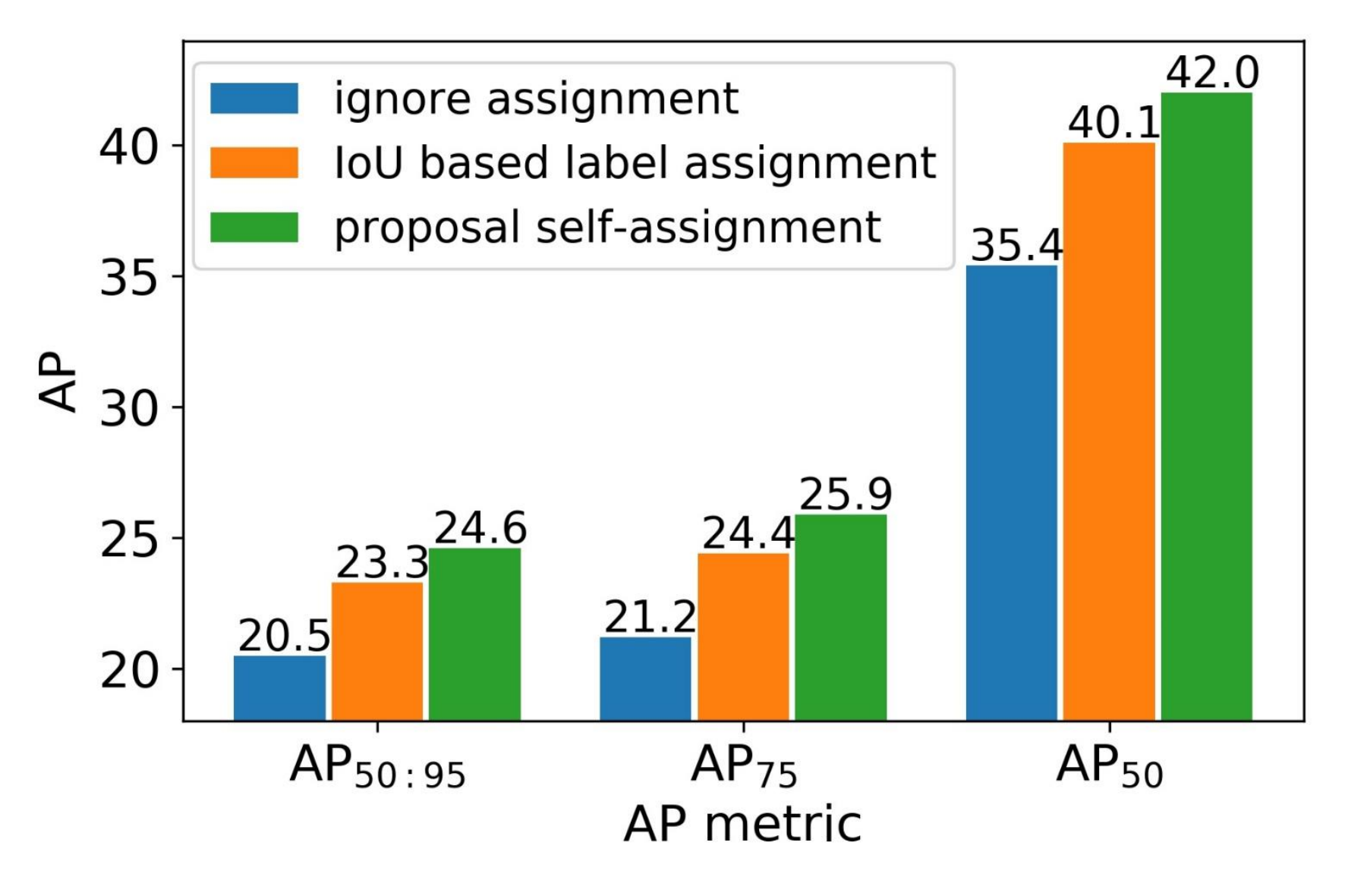}
    \caption{}
    \label{fig:6-c}
  \end{subfigure}
    \vskip -0.1in
  \caption{Ablation study about: (a) the updating interval of ACT. (b) the effect of reliable ratio $\alpha$. (c) different label assignment strategies.}
  \label{fig:6}
    \vskip -0.1in
\end{figure*}

\vspace{3mm}
\noindent
{\bf The necessity of ACT adaptation.} We randomly select a subset of unlabeled data to determine ACT for every $K$ iterations in our implementation. As mentioned in \cref{subsec: labelmatch}, the proposed ACT are updated to the evolved teacher during the training phase, avoiding a false bias caused by the outdated predictions. \cref{fig:6-a} proves the necessity of ACT adaptation, where the performance is pretty bad without adaptation ($K=+\infty$). In our experiments, we simply use $K=1000$ and a subset of unlabeled data (10,000) to refresh the thresholds.

\vspace{2mm}
\noindent
{\bf Effect of reliable ratio $\alpha$.} In the training phase, we split the candidate pseudo labels into reliable pseudo labels and uncertain pseudo labels by $\alpha\%$ according to the confidence scores. Here, we analyze the influence of different ratios. As shown in \cref{fig:6-b}, if we directly set all the candidate pseudo labels as uncertain pseudo labels ($\alpha=0$), the performance is worse than splitting some pseudo labels to reliable, which is mainly caused by the lack of box regression optimization for the unlabeled data. However, setting too many pseudo labels as reliable pseudo labels is also harmful due to the noisy boxes. We use $\alpha=20$ for all experiments.

\begin{figure}\centering
  \begin{tabular}{ *{2}{c @{\hspace{1\tabcolsep}} c @{\hspace{1\tabcolsep}} c} }
     \rotatebox{90}{\hspace{0.5cm}\small{IoU-based}} & 
      \includegraphics[width=0.45\linewidth]{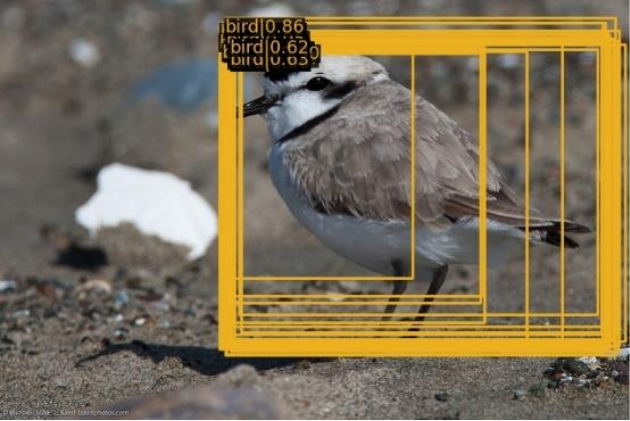} &
      \includegraphics[width=0.45\linewidth]{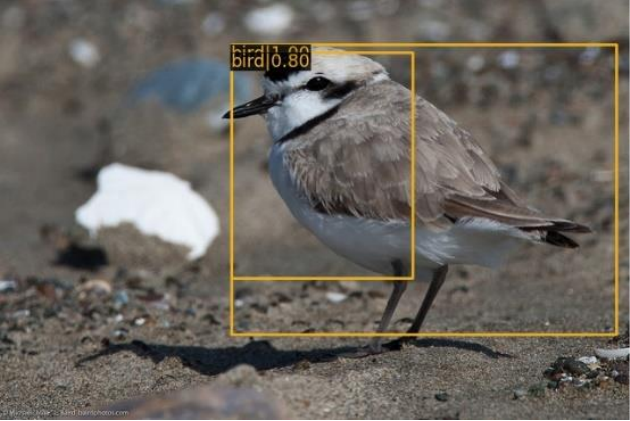} \\
      \rotatebox{90}{\hspace{0.6cm}\small{proposal}} & 
        \includegraphics[width=0.45\linewidth]{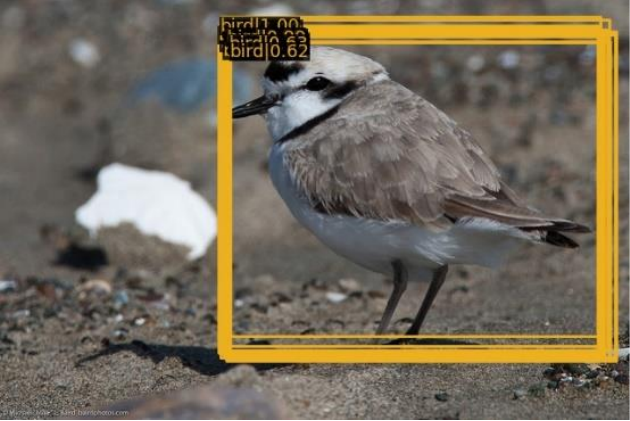} &
        \includegraphics[width=0.45\linewidth]{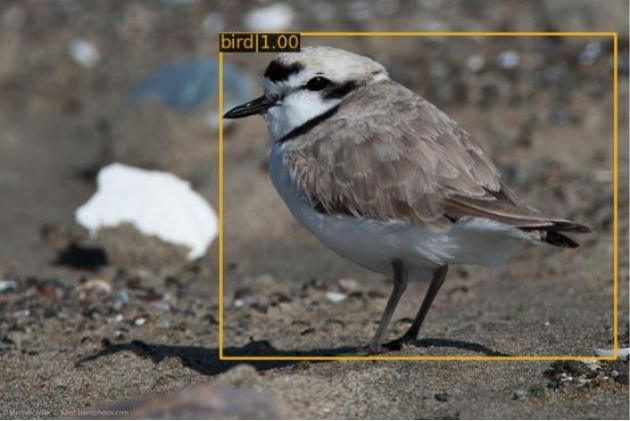} \\
        & \small{before NMS} & \small{after NMS}
        \end{tabular}
    \vskip -0.1in
    \caption{Visualization of the predictions before and after NMS post-processing. IoU based: the model is trained based on IoU based label assignment; proposal: the model is trained based on proposal self-assignment}
\label{fig:7}
  \vskip -0.1in
\end{figure}

\vspace{2mm}
\noindent
{\bf Proposal self-assignment.} We compare different label assignment strategies for uncertain pseudo labels: 1) Ignore assignment; 2) IoU based label assignment; 3) Proposal self-assignment. Here the ignore assignment means that the uncertain pseudo labels are directly set as ignore labels. \cref{fig:6-c} shows the superiority of the proposal self-assignment strategy over other label assignments. For the ignore assignment, there exists an imbalance between foreground and background, and background dominates the object detection training, which makes the ignored objects tend to be regarded as background after training, impeding performance improvement. For IoU based label assignment, it will produce many ambiguous boxes after training due to the instance mismatch as illustrated in \cref{fig:7}.

\begin{figure}
  \centering
  \begin{subfigure}{0.52\linewidth}
    \includegraphics[width=\linewidth]{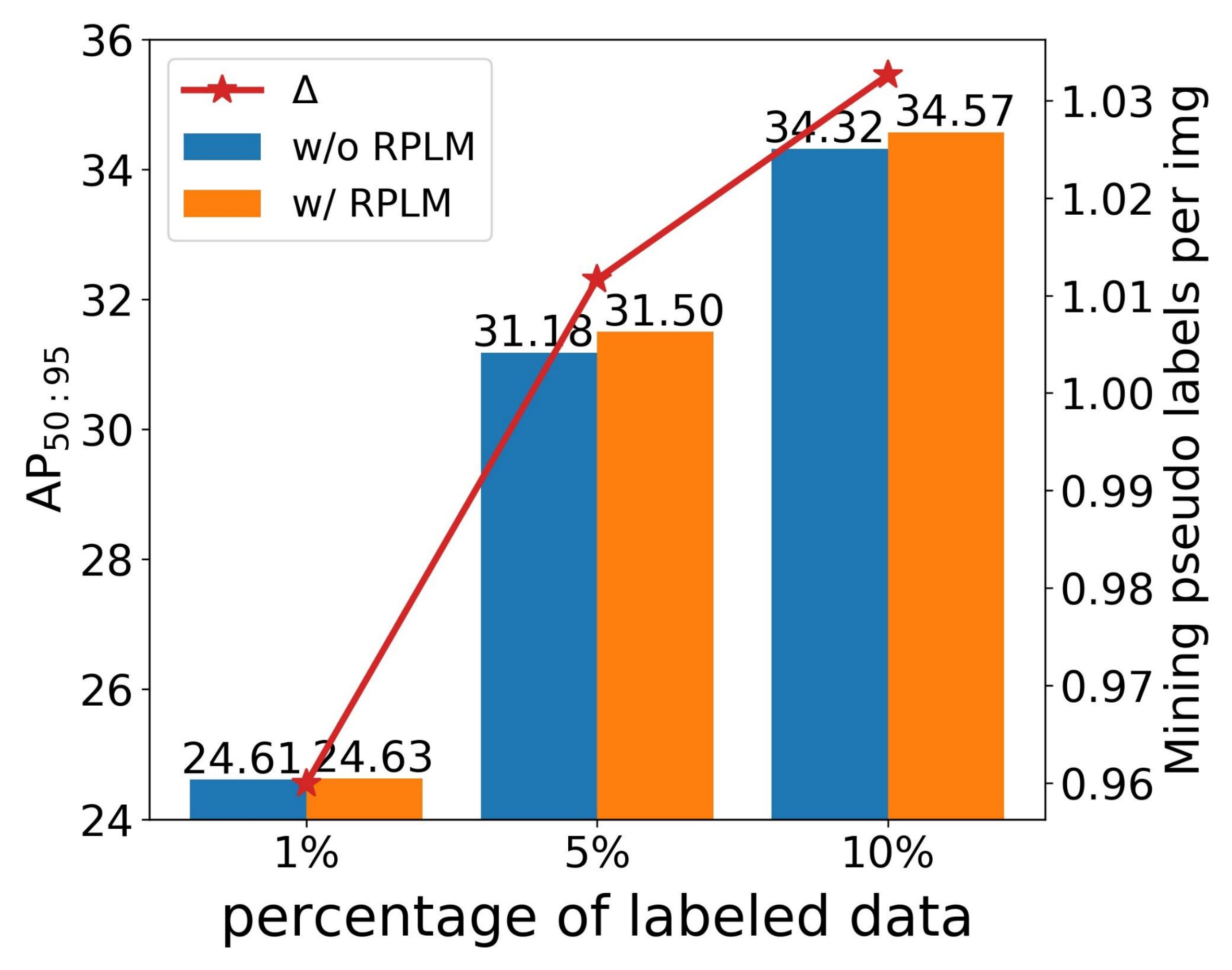}
    \caption{}
    \label{fig:8-a}
  \end{subfigure}
    \begin{subfigure}{0.47\linewidth}
    \includegraphics[width=\linewidth]{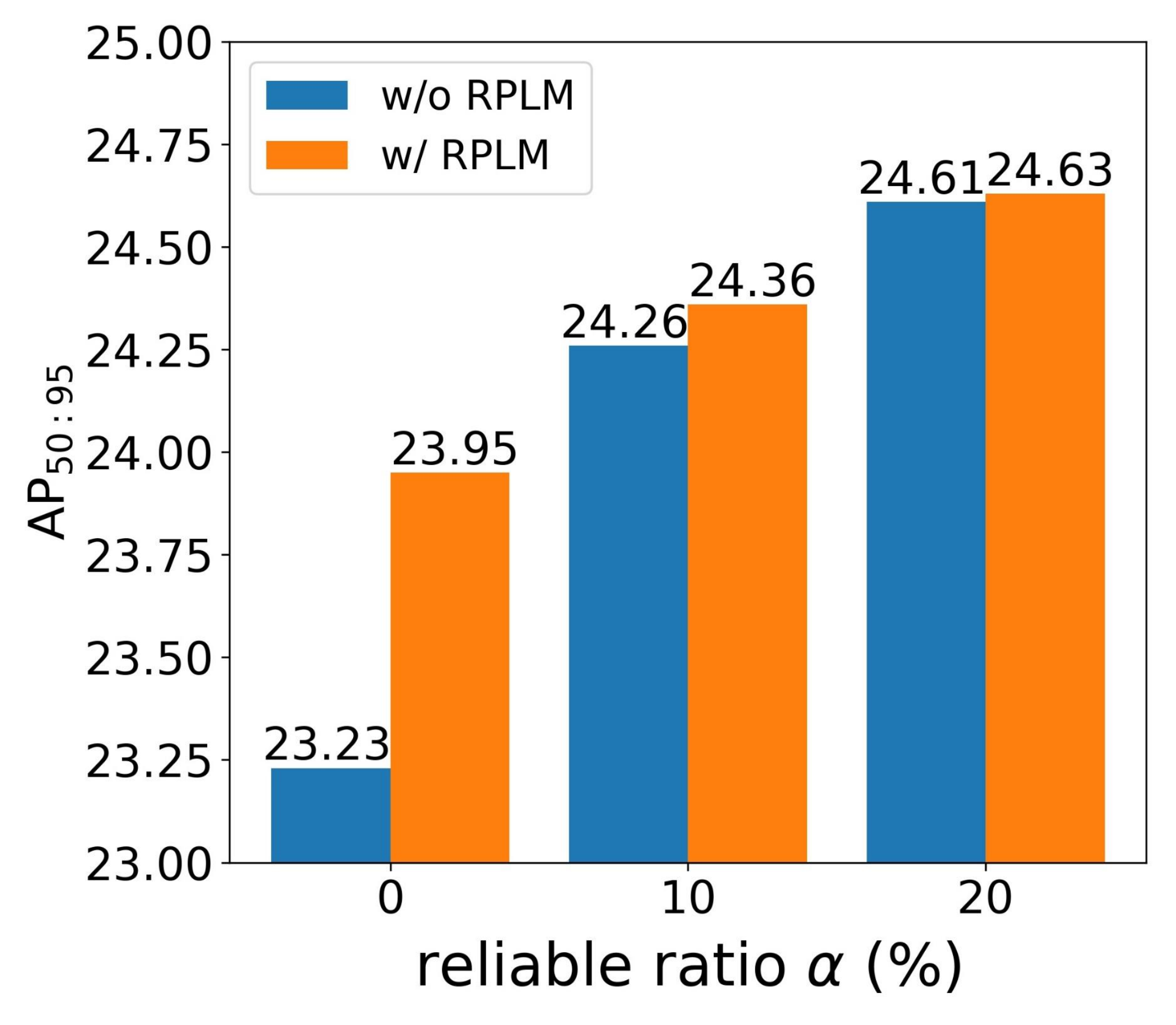}
    \caption{}
    \label{fig:8-b}
  \end{subfigure}
    \vskip -0.1in
  \caption{Ablation study about RPLM. (a) RPLM in different percentages of labeled data. ($\Delta$ means the number of mining pseudo labels per image.) (b) RPLM in different reliable ratio $\alpha$.}
  \label{fig:8}
    \vskip -0.1in
\end{figure}

\vspace{2mm}
\noindent
{\bf Effect of RPLM.} To verify the effectiveness of RPLM, we estimate it in the setting of COCO-standard. As depicted in \cref{fig:8-a}, RPLM slightly boosts the performance, but remitting the sensitivity of the reliable ratio $\alpha$ and showing an obvious improvement even $\alpha$ dropping to zero (\cref{fig:8-b}). This implies that, with the favor of RPLM, we can easily adjust $\alpha$ to make stable performance improvement without expert techniques.

\section{Conclusion and Future Work}
\label{conclusion}

In this paper, we first diagnose the existing SSOD frameworks experimentally and figure out the common limitation, namely label mismatch problem, including two different but complementary granularities, \emph{i.e.}, distribution-level and instance-level. To solve the above problems, we propose the LabelMatch framework. For distribution-level mismatch, LabelMatch develops a re-distribution mean teacher to derive adaptive label-distribution-aware confidence thresholds by narrowing the class distribution discrepancy between labeled and unlabeled data, and then generating unbiased pseudo labels. For instance-level mismatch, LabelMatch adopts a proposal self-assignment method, injecting the proposals generated by the student into the teacher model to supervise proposals correction. Furthermore, a reliable pseudo label mining strategy is introduced to convert high-quality uncertain pseudo labels to reliable ones, facilitating the cycle of positive feedback during self-training. Extensive experimental results verify the efficacy of the proposed LabelMatch, which establishes a new state-of-the-art on both PASCAL-VOC and MS-COCO datasets.

\noindent\textbf{Limitations.} While we have shown the superiority of LabelMatch, there is still a non-negligible problem that the labeled and unlabeled data are assumed to follow the same distribution. Hence, LabelMatch relies on the class distribution prior, which is inaccessible in some scenarios, \emph{e.g.}, unsupervised domain adaptive object detection. It is beneficial to study the label mismatch problems between two different distributions and, intuitively, advance the LabelMatch to more complex situations without class distribution prior, which is an interesting future work.

\section*{Acknowledgements} 
This work was sponsored in part by National Natural Science Foundation of China (U20B2066, 62106220), and Hikvision Open Fund.

{\small
\bibliographystyle{ieee_fullname}
\bibliography{egbib}
}

\appendix

\section{Consistent Class Distribution Assumption}
\begin{figure}[t]
  \centering
  \begin{subfigure}{0.495\linewidth}
    \includegraphics[width=\linewidth]{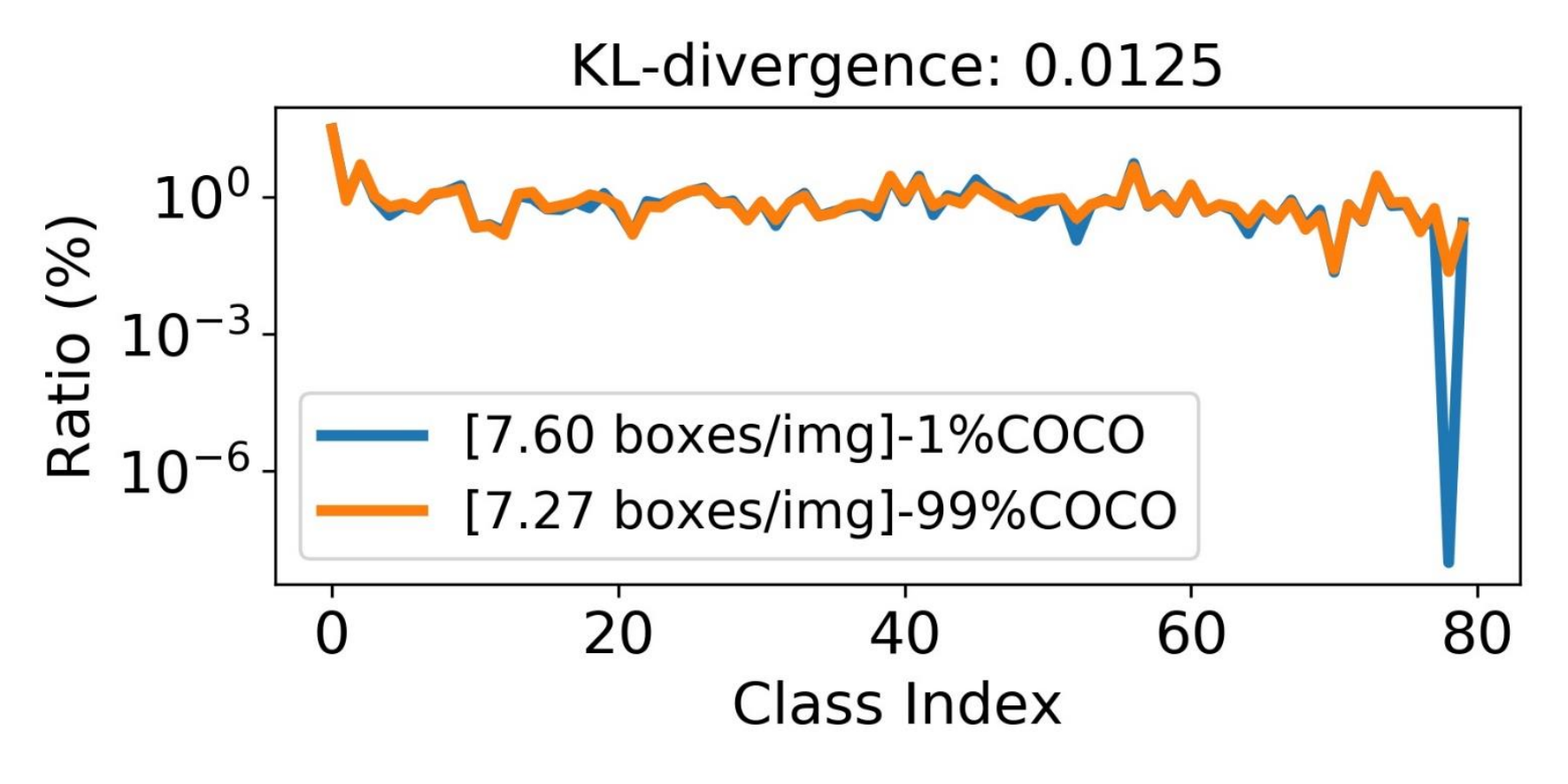}
  \end{subfigure}
  \hfill
  \begin{subfigure}{0.495\linewidth}
    \includegraphics[width=\linewidth]{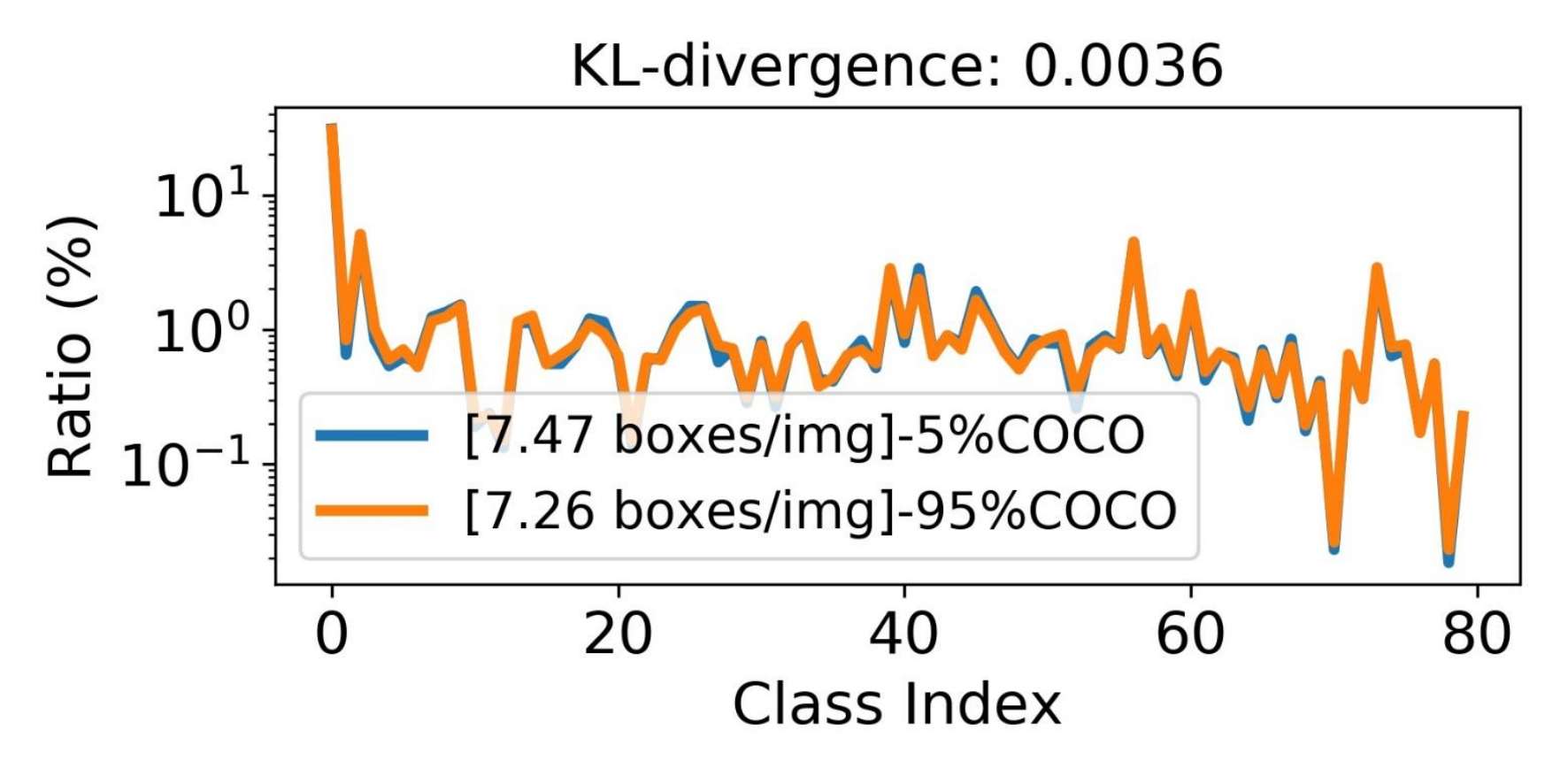}
  \end{subfigure}
  \begin{subfigure}{0.495\linewidth}
    \includegraphics[width=\linewidth]{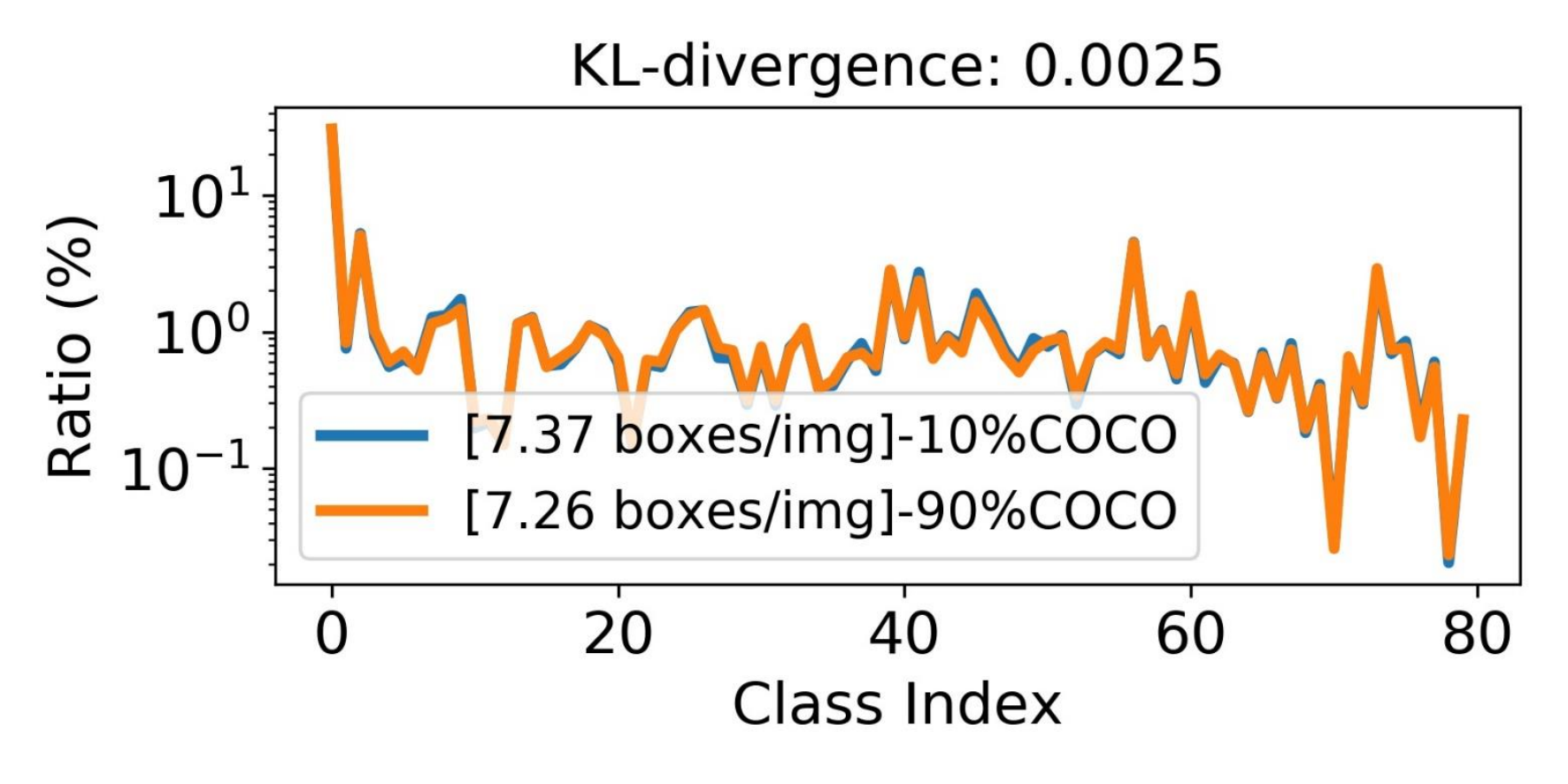}
  \end{subfigure}
  \hfill
  \begin{subfigure}{0.495\linewidth}
    \includegraphics[width=\linewidth]{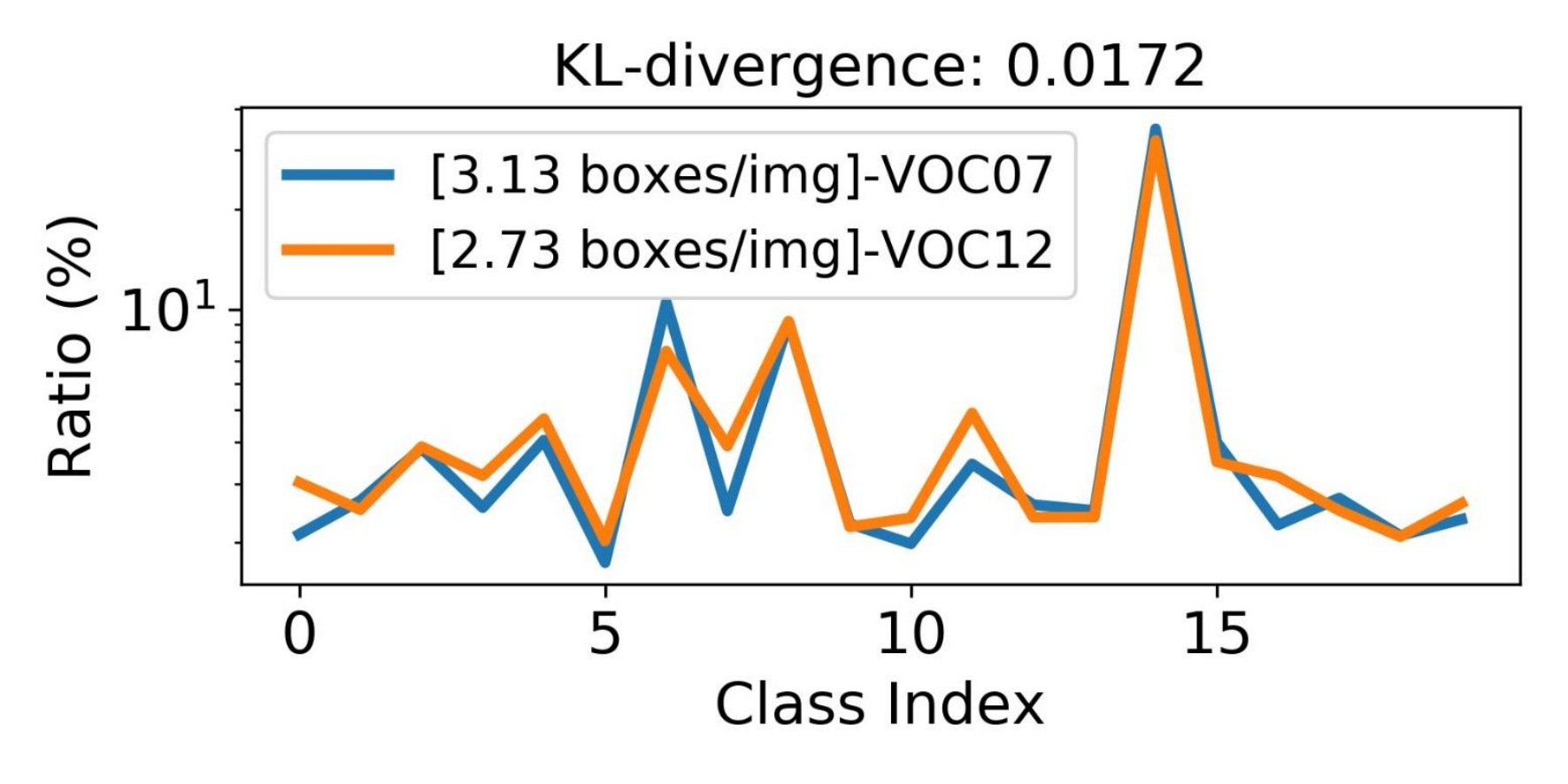}
  \end{subfigure}
  \caption{Comparisons on class distribution between the labeled and unlabeled data. The blue and orange lines denote the foreground-foreground class distribution in the labeled and unlabeled data, respectively. ``boxes/img'' in the legend represents the foreground-background ratio.}
  \label{fig:a1}
\end{figure}

LabelMatch is based on the assumption that consistent class distribution exists between the labeled and unlabeled data since they are drawn from the same data distribution. To further verify this hypothesis, we present the comparisons between the labeled and unlabeled data in COCO-standard and VOC using the ground-truth labels. As shown in \cref{fig:a1}, the foreground-foreground class distribution and the foreground-background ratio of the unlabeled data are close to those of the labeled data in these SSOD settings. 

\section{More Results on COCO-standard}
In this section, we present more experimental results on COCO-standard using the ablation study setting (see the fifth column in \cref{tab:a7}).
Firstly, we carry out more analysis about ACT in \cref{subsec: act}. Then, we study the effect of hyper-parameter in RPLM in \cref{subsec: rplm} and more analysis about proposal self-assignment in \cref{subsec: proposal}. Finally, more qualitative results are exhibited in \cref{subsec: visual}.

\subsection{Analysis of ACT}
\label{subsec: act}
In this part, we present more analysis about the proposed ACT from flexibility and implementation.

\begin{figure}[t]
  \centering
  \begin{subfigure}{0.495\linewidth}
    \includegraphics[width=\linewidth]{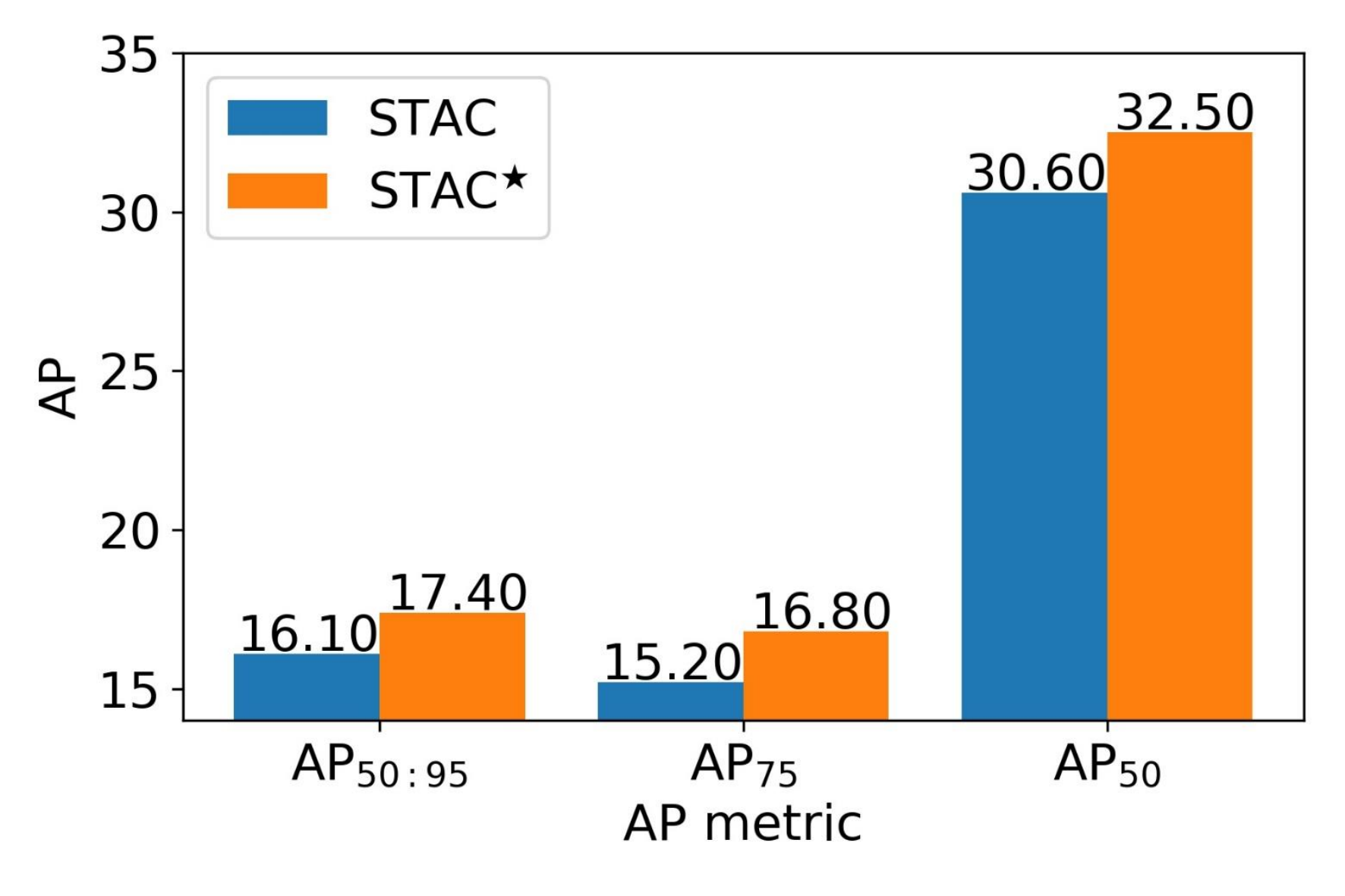}
  \end{subfigure}
  \hfill
  \begin{subfigure}{0.495\linewidth}
    \includegraphics[width=\linewidth]{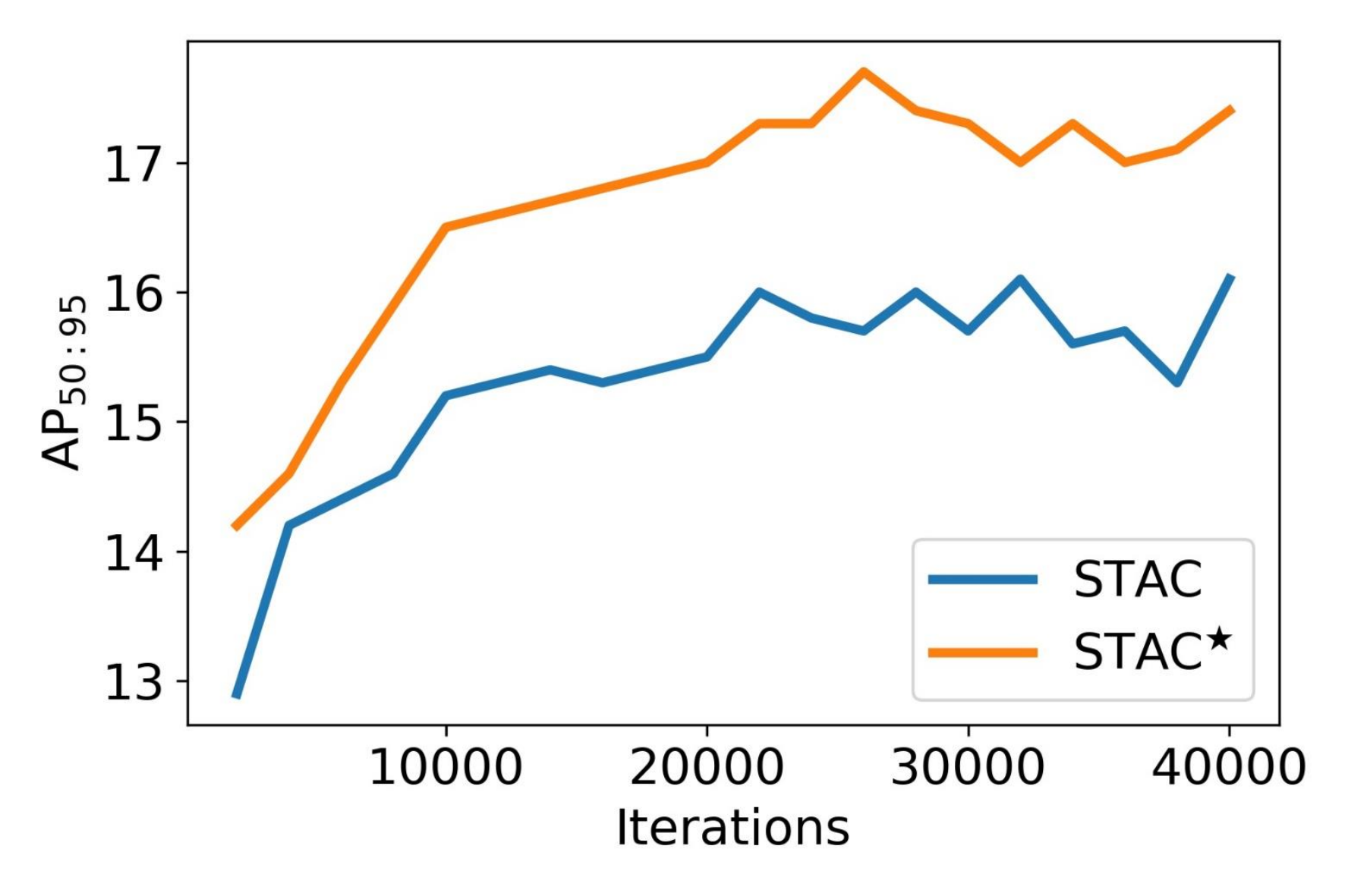}
  \end{subfigure}
  \caption{Performance comparisons between STAC and STAC$^\star$ on COCO-standard with 1\% labeled data. }
  \label{fig:a2}
\end{figure}

\vspace{1mm}
\noindent
{\bf Flexibility.} To further demonstrate the flexibility of our method, we extend STAC~\cite{sohn2020a} with the proposed ACT, denoted as STAC$^\star$ for short. The original STAC first uses a pretrained model to generate pseudo labels and then uses a threshold of 0.9 to filter out low-quality pseudo labels, which are finally fed back into the network with strong data augmentation for model fine-tuning. Alternatively, STAC$^\star$ replaces the fixed threshold with the proposed ACT for pseudo labeling and updates the thresholds every epoch. Since there is no mean teacher in STAC, the label assignment strategy of STAC$^\star$ simply follows the \emph{ignore assignment}, where uncertain pseudo labels are set as ignore labels. As shown in \cref{fig:a2}, there is an apparent performance gain after equipping STAC with ACT, demonstrating the universality of the proposed ACT.

\begin{table}[t]
  \centering
  \resizebox{0.48\textwidth}{!}{
    \begin{tabular}{>{\centering}p{0.1\textwidth}>{\centering}p{0.05\textwidth}>{\centering}p{0.08\textwidth}>{\centering}p{0.05\textwidth}>{\centering}p{0.05\textwidth}>{\centering\arraybackslash}p{0.05\textwidth}}
    \toprule
     & ACT & iterations & 1\% & 5\% & 10\% \\
    \midrule
    LabelMatch & Offline & 40K & 24.6 & 31.6 & 34.6 \\
    LabelMatch & Online & 40K & 24.6 & 31.5 & 34.6 \\
    \bottomrule
  \end{tabular}}
  \caption{Performance ($AP_{50:95}$) comparisons between the online and offline versions of ACT. We only run 1-fold using the ablation training setting.}
  \label{tab:a2}
\end{table}

\begin{figure}[t]
  \centering
      \begin{subfigure}{0.48\linewidth}
      \includegraphics[width=\linewidth]{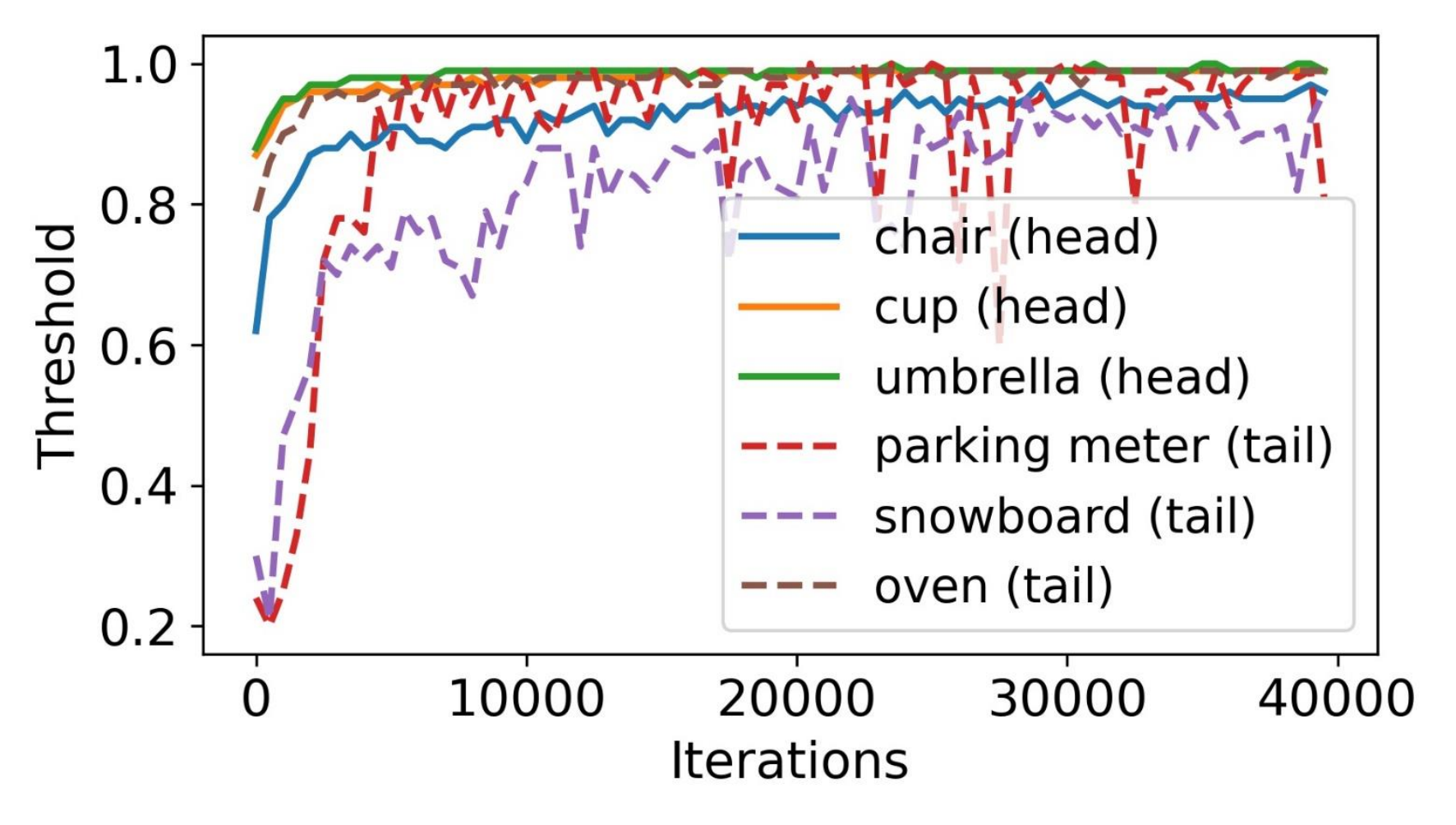}
      \caption{}
      \label{fig:2-a}
      \end{subfigure}
  \hfill
      \begin{subfigure}{0.48\linewidth}
      \includegraphics[width=\linewidth]{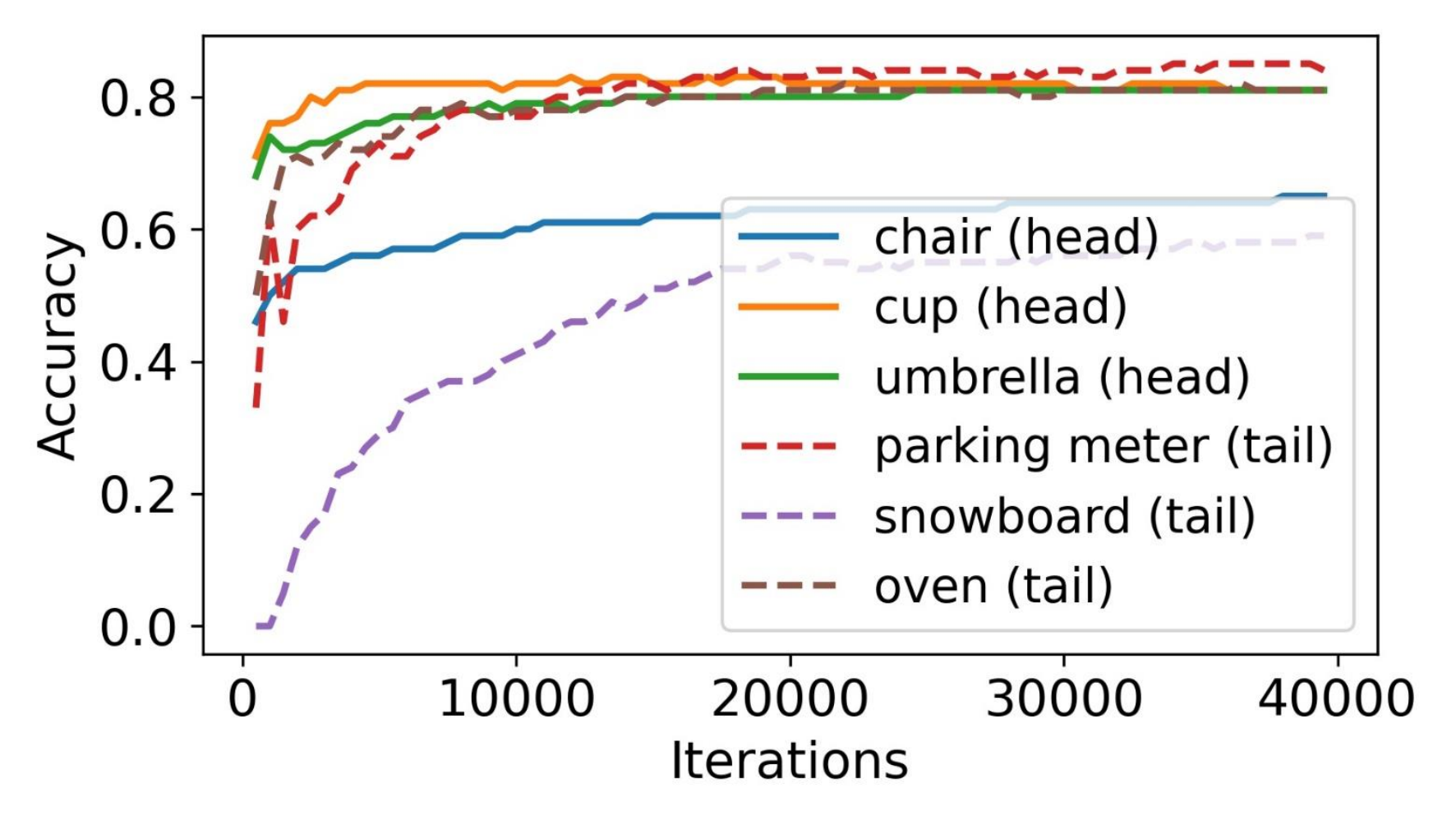}
      \caption{}
      \label{fig:2-c}
      \end{subfigure}
  \caption{(a) Thresholds in the training phrase. (b) The quality of reliable pseudo labels.}
  \label{fig: b2}
     \vskip -0.15in
\end{figure}

\begin{figure*}[t]
  \centering
  \begin{subfigure}{0.765\linewidth}
    \includegraphics[width=\linewidth]{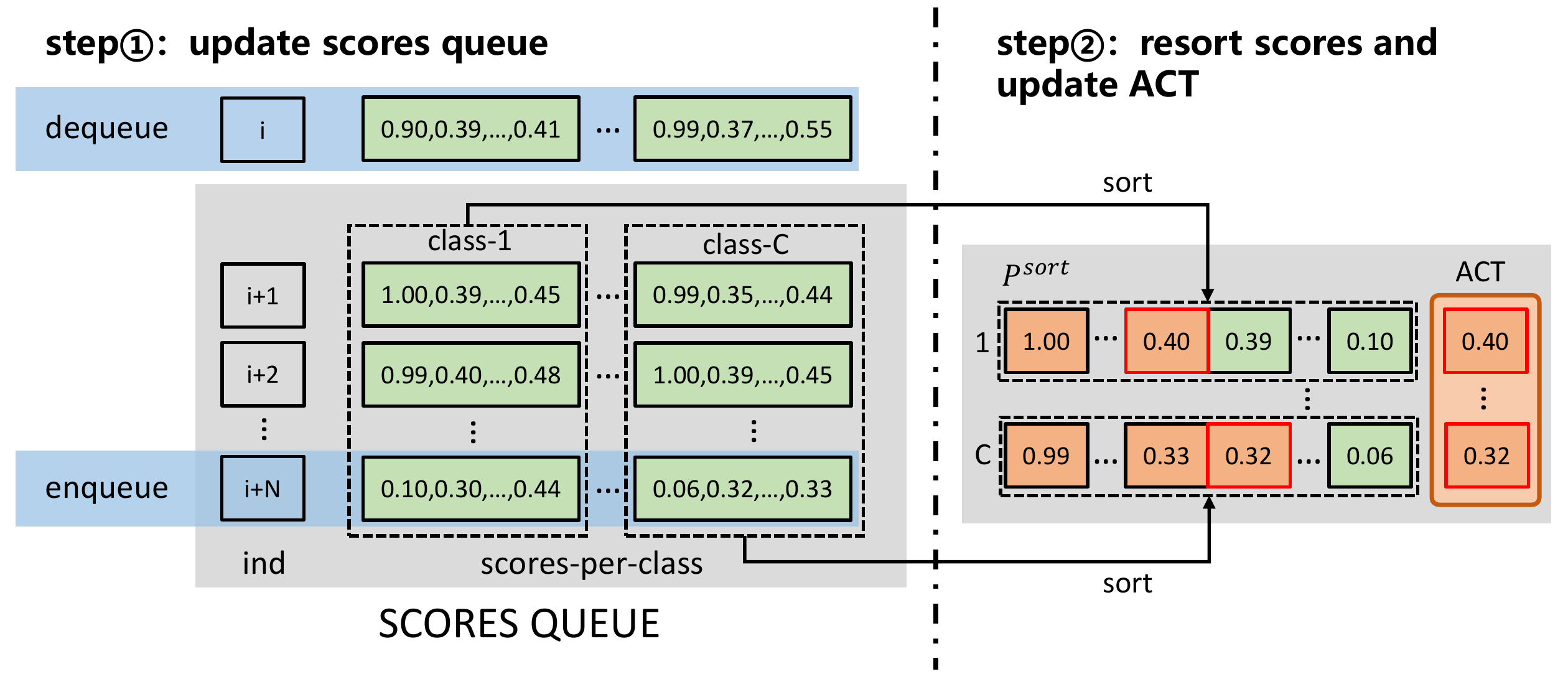}
    \caption{}
    \label{fig:a3-a}
  \end{subfigure}
  \hfill
  \begin{subfigure}{0.225\linewidth}
    \includegraphics[width=\linewidth]{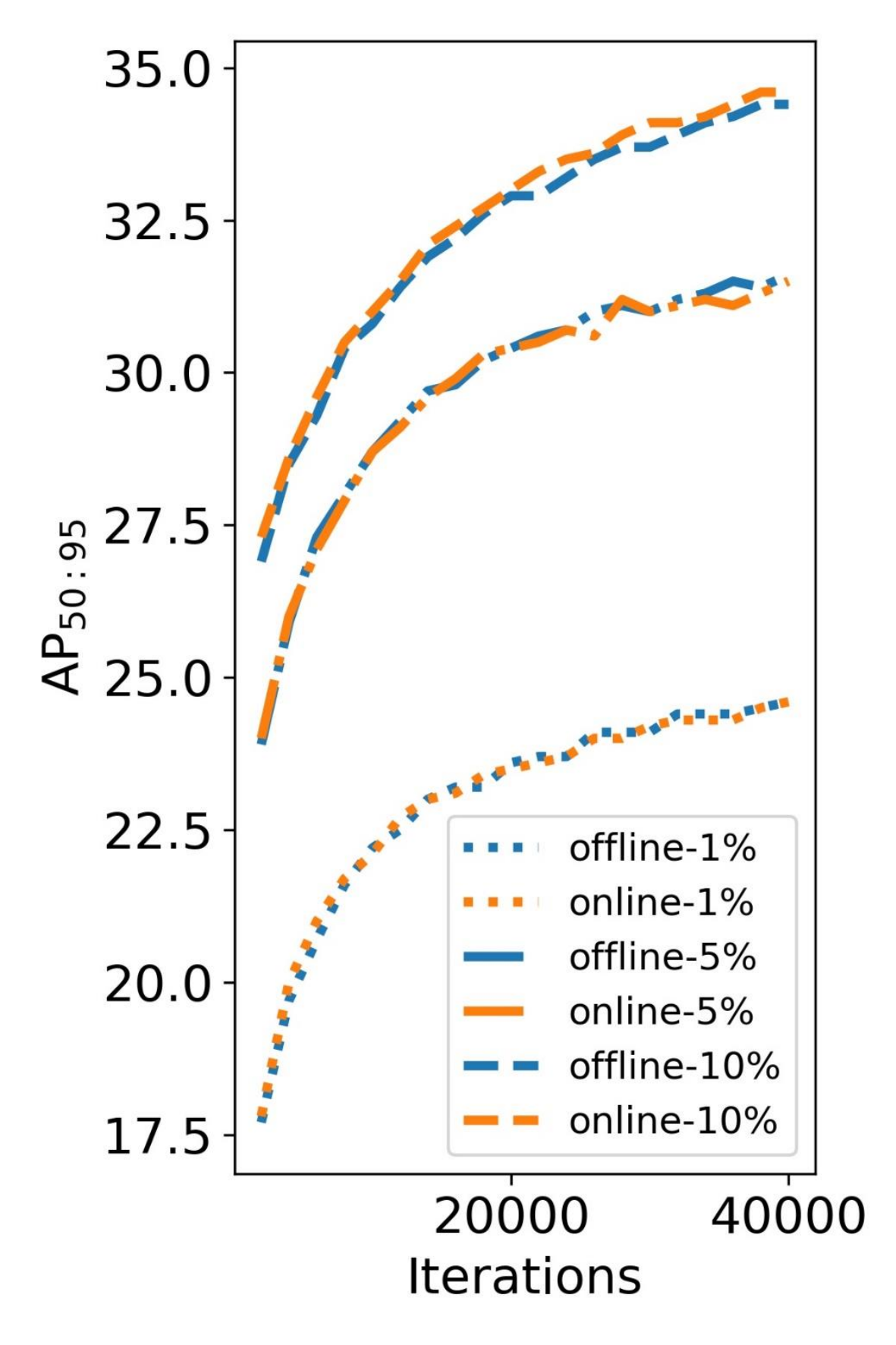}
    \caption{}
    \label{fig:a3-b}
  \end{subfigure}
  \caption{(a) Implementation of the online version of ACT. Each training iteration consists of two steps: (1) obtain predictions from the teacher model and update the scores queue; (2) re-sort scores and update the ACT in real-time. (b) Performance comparison between the online and offline version of ACT during the training phase on three COCO-standard settings with 1\%, 5\% and 10\% labeled data.}
  \label{fig:a3}
\end{figure*}

\vspace{1mm}
\noindent
{\bf Online vs. Offline.} As discussed in the paper, ACT are updated to the evolved teacher during the training phase, avoiding a negative bias caused by the outdated predictions. There are two patterns to update ACT, one of which is introduced in the paper, leveraging a subset of unlabeled data to update ACT every $K$ iterations, termed as offline version. Here, we describe another pattern, named as online version, which maintains a scores queue, as shown in \cref{fig:a3-a}. The teacher's prediction is pushed into the scores queue for refreshing the ACT in each training iteration, which can be seen as a special case of the offline version with $K=1$. Both versions of ACT can get satisfactory performance, as shown in \cref{fig:a3-b} and \cref{tab:a2}. We use the offline version in all the experiments and will release the online version as well.

\vspace{1mm}
\noindent
{\bf Thresholds evolve alone training.} We select three head classes and three tail classes on offline version for analysis. As shown in \cref{fig: b2}, the thresholds (Eq.6 in the paper to filter reliable pseudo labels) increase in both head and tail classes during optimization. Specifically, the thresholds for tail classes are more fluctuated than those for head classes due to the scarce samples. Also, the quality of reliable pseudo labels get increased as training goes on.

\subsection{Analysis of RPLM Hyper-Parameter}
\label{subsec: rplm}

There are two hyper-parameters ($T_{score}, T_{iou}$) in the component of reliable pseudo label mining (RPLM). Here we use COCO-standard with 10\% labeled data as the experimental setting. As shown in \cref{tab:a1}, the best performance appears when $(T_{score}, T_{iou})=(0.8, 0.8)$. Therefore, we use $(T_{score}, T_{iou})=(0.8, 0.8)$ by default in all experiments throughout the paper. It is also worth mentioning that our method is not sensitive to these hyper-parameters.

\subsection{Analysis of proposal self-assignment}
\label{subsec: proposal}

\begin{figure}[t]
  \centering
      \begin{subfigure}{0.48\linewidth}
      \includegraphics[width=\linewidth]{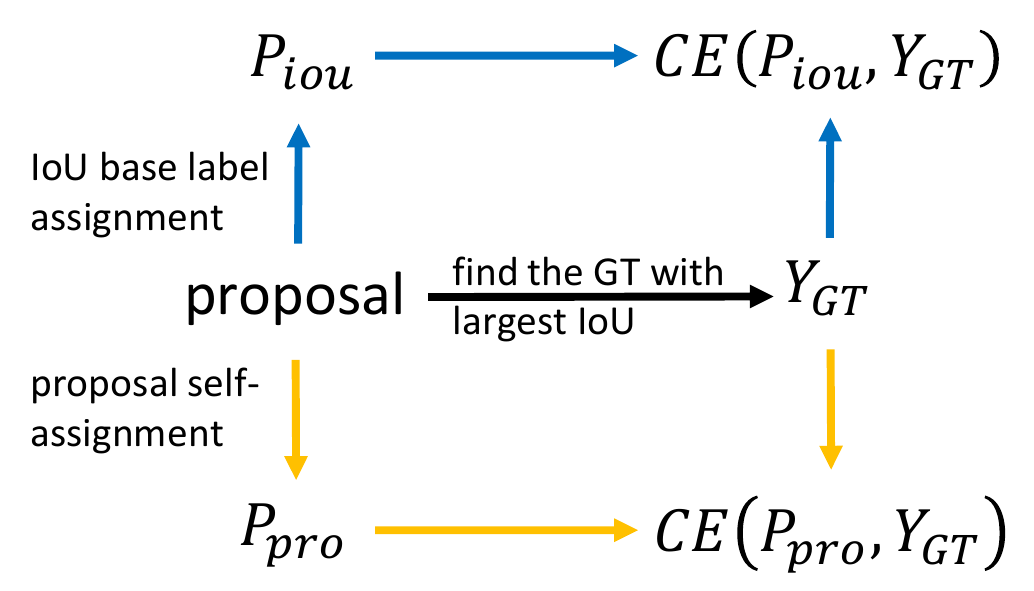}
      \caption{}
      \label{fig:1-a}
      \end{subfigure}
  \hfill
      \begin{subfigure}{0.48\linewidth}
      \includegraphics[width=\linewidth]{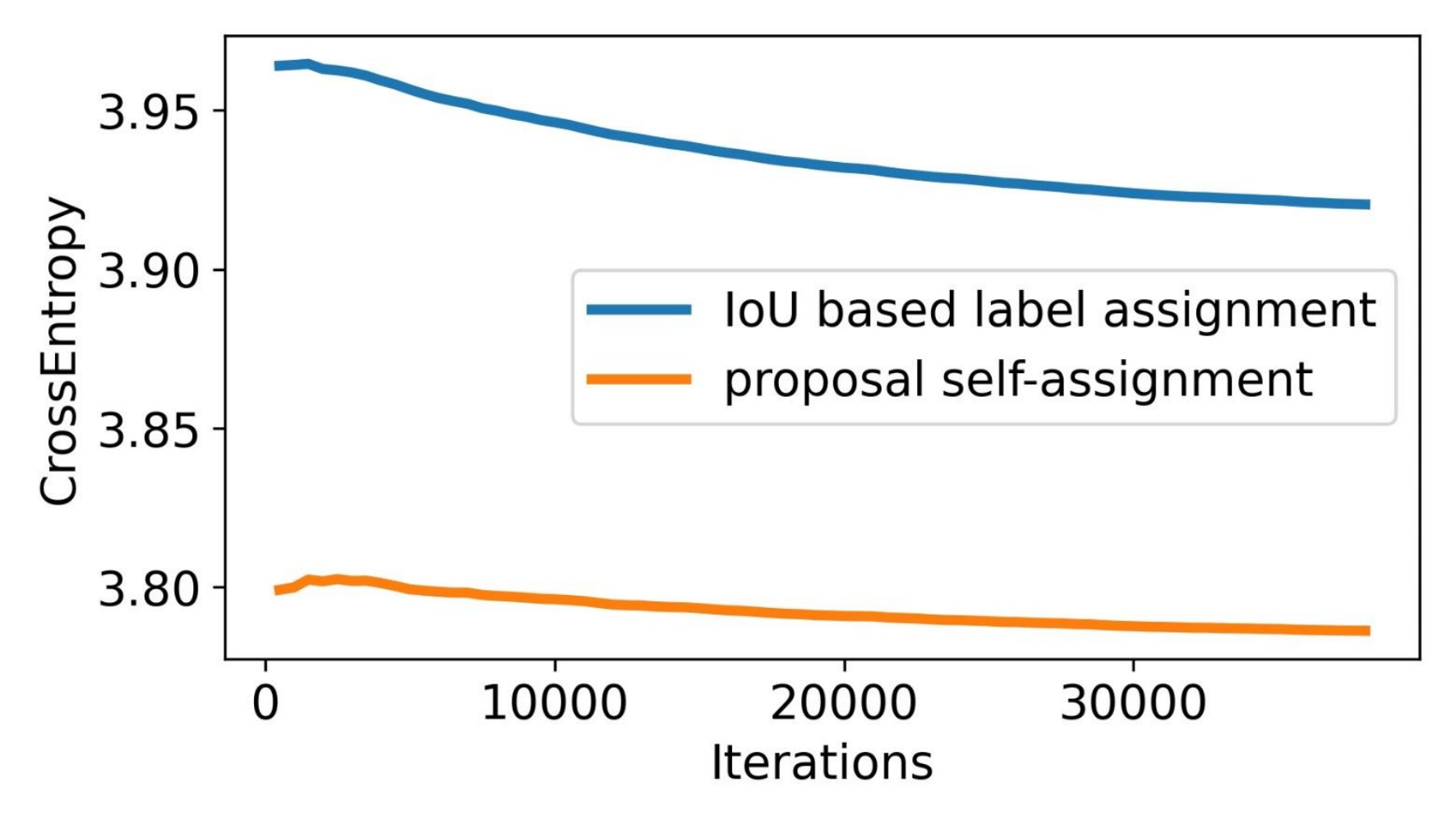}
      \caption{}
      \label{fig:1-b}
      \end{subfigure}
  \caption{{\small (a) The solution to measure the quality of predictions. (b) The quality comparison of the predictions between two label assignment methods in the training phase (lower is better).}}
  \label{fig: b1}
     \vskip -0.04in
\end{figure}

To further analyze the quality of the teacher's RoI head predictions on the student's proposals (proposals self-assignment \emph{vs.} IoU-based label assignment), we use the ground truth for quantitative measurement. For each proposal, we calculate the cross-entropy between the corresponding prediction and the nearest ground truth (set as background if IoU$<$0.5). As shown in \cref{fig: b1}, the predictions by proposal self-assignment show better quality than IoU-based one.

\begin{table}[t]
  \centering
  \resizebox{0.48\textwidth}{!}{
  \begin{tabular}{cc|cc|cc}
    \toprule
    ($T_{score}$, $T_{iou}$) & $AP_{50:95}$ & ($T_{score}$, $T_{iou}$) & $AP_{50:95}$ & ($T_{score}$, $T_{iou}$) & $AP_{50:95}$ \\
    \midrule
    (0.7, 0.7) & 34.4 & (0.8, 0.7) & 34.4 & (0.9, 0.7) & 34.4 \\
    (0.7, 0.8) & 34.5 & (0.8, 0.8) & \textbf{34.6} & (0.9, 0.8) & 34.5 \\
    (0.7, 0.9) & 34.5 & (0.8, 0.9) & \textbf{34.6} & (0.9, 0.9) & 34.3 \\
    \bottomrule
  \end{tabular}}
  \caption{Effect of hyper-parameters in RPLM.}
  \label{tab:a1}
    \vskip -0.05in
\end{table}

\begin{figure*}[t]
  \centering
    \includegraphics[width=\linewidth]{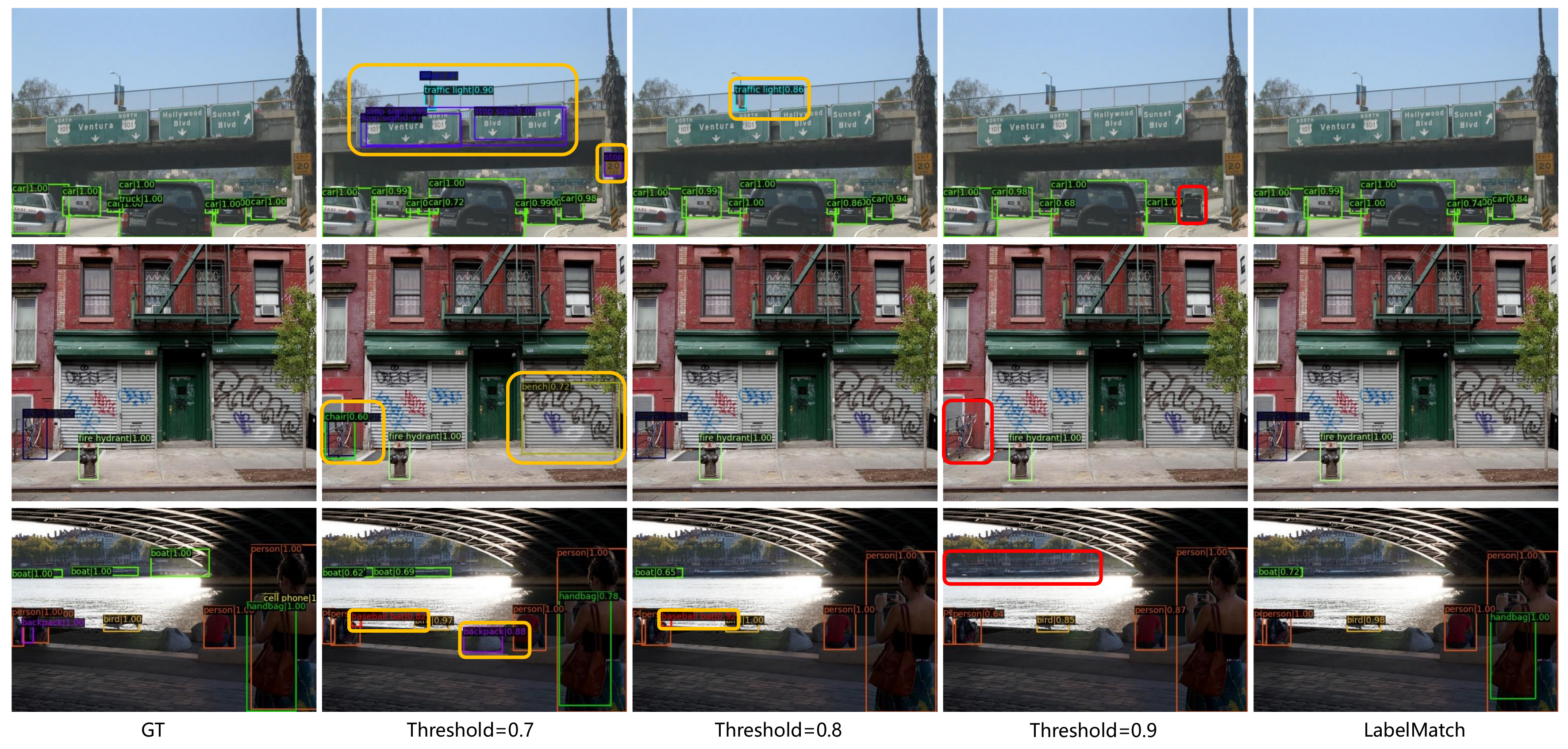}
  \caption{Qualitative comparisons between the single confidence threshold and the proposed LabelMatch. Red rectangles highlight the false negatives, and yellow rectangles highlight the false positives. The score threshold for visualization is 0.6.}
  \label{fig:a4}
\end{figure*}

\begin{table*}[t]
  \centering
  \resizebox{0.98\textwidth}{!}{
    \begin{tabular}{>{\centering}p{0.18\textwidth}>{\centering}p{0.2\textwidth}>{\centering}p{0.2\textwidth}>{\centering}p{0.2\textwidth}>{\centering\arraybackslash}p{0.2\textwidth}}
    \toprule
     Data Split& Normal$\to$Foggy & Small$\to$Large & Across cameras & Synthetic$\to$Real \\
    \midrule
    labeled data & Cityscapes (train) & Cityscapes (train) & KITTI & Sim10K \\ 
    unlabeled data & Cityscapes-foggy (train) & BDD100K (train) & Cityscapes (train) & Cityscapes (train) \\ 
    test data & Cityscapes-foggy (val) & BDD100K (val) &  Cityscapes (val) & Cityscapes (val) \\ 
    \bottomrule
  \end{tabular}}
  \caption{Four differnt domain shifts in DA-OD, which are contructed by five different datasets, including Cityscapes~\cite{cordts2016the}, Cityscapes-foggy~\cite{sakaridis2018semantic}, KITTI~\cite{sakaridis2018semantic}, Sim10k~ \cite{johnson-roberson2017driving} and BDD100K~\cite{johnson-roberson2017driving}.}
  \label{tab:a3}
    \vskip -0.1in
\end{table*}

\subsection{Qualitative Results}
\label{subsec: visual}
We perform the qualitative comparisons between the proposed method and the mean teacher frameworks with a fixed and single confidence threshold (varying from 0.7 to 0.9). As shown in \cref{fig:a4}, there are many false positives with a low confidence threshold (yellow rectangles in the second column), while many false negatives appear when using a high confidence threshold (red rectangles in the fourth column). Although the manual search threshold (0.8) via trial-and-error can achieve satisfactory results, our method shows even better qualitative results.

\section{Domain Adaptive Object Detection}
\label{sec: uda}
LabelMatch is based on the consistent class distribution assumption between the labeled and unlabeled data. To explore the robustness of LabelMatch to the prior dependence on this assumption, we extend it to the scenario of domain adaptive object detection (DA-OD) \cite{saito2019strong,xu2020cross,vs2021mega,SFOD} where the labeled source data and the unlabeled target are drawn from two different data distributions.

\begin{table}[t]
\centering
\resizebox{0.48\textwidth}{!}{
\begin{tabular}{l|cccccccc|c}
\toprule
 Method& \rotatebox{90}{truck} & \rotatebox{90}{car} & \rotatebox{90}{rider} & \rotatebox{90}{person} & \rotatebox{90}{train} & \rotatebox{90}{motor} & \rotatebox{90}{bicycle} & \rotatebox{90}{bus} & mean \\
 \midrule
 Source only & 19.2 & 47.9 & 40.8 & 34.8 & 7.8 & 24.2 & 36.0 & 36.4 & 30.9 \\ \hline
 CVPR2020:GPA \cite{xu2020cross} & 24.7 & 54.1 & 46.7 & 32.9 & 41.1 & 32.4 & 38.7 & 45.7 & 39.5 \\
 CVPR2020:HTCN \cite{chen2020harmonizing} & 31.6 & 47.9 & 47.5 & 33.2 & 40.9 & 32.3 & 37.1 & 47.4 & 39.8 \\
 CVPR2021:MeGA \cite{vs2021mega} & 25.4 & 52.4 & 49.0 & 37.7 & 46.9 & 34.5 & 39.0 & 49.2 & 41.8 \\
 CVPR2021:UMT \cite{deng2021unbiased} & 34.1 & 48.6 & 46.7 & 33.0 & 46.8 & 30.4 & 37.3 & 56.5 & 41.7 \\ \hline
 LabelMatch (Ours) & \textbf{42.0} & \textbf{62.2} & \textbf{55.4} & \textbf{45.3} & \textbf{55.1} & \textbf{43.5} & \textbf{51.5} & \textbf{64.1} & \textbf{52.4} \\
\bottomrule
\end{tabular}}
\caption{Results of adaptation from normal to foggy weathers. ``Source only'' refers to the model trained by labeled source data.}
\label{tab: a4-1}
\vskip -0.05in
\end{table}

\begin{table}[t]
\centering
\resizebox{0.48\textwidth}{!}{
\begin{tabular}{l|cccccccc|c}
\toprule
 Method& \rotatebox{90}{truck} & \rotatebox{90}{car} & \rotatebox{90}{rider} & \rotatebox{90}{person} & \rotatebox{90}{train} & \rotatebox{90}{motor} & \rotatebox{90}{bicycle} & \rotatebox{90}{bus} & mean \\
 \midrule
 Source only & 18.3 & 50.0 & 33.3 & 35.8 & - & 18.4 & 27.6 & 17.0 & 28.7 \\ \hline
 CVPR2019:SW-Faster \cite{saito2019strong} & 15.2 & 45.7 & 29.5 & 30.2 & - & 17.1 & 21.2 & 18.4 & 25.3 \\
 CVPR2020:CR-DA \cite{xu2020exploring} & 19.5 & 46.3 & 31.3 & 31.4 & - & 17.3 & 23.8 & 18.9 & 26.9 \\ \hline
 LabelMatch (Ours) & 39.4 & 54.6 & 37.4 & 42.9 & - & 25.7 & 29.8 & 41.7 & 38.8 \\
 LabelMatch$^\dagger$ (Ours) & \textbf{39.8} & \textbf{55.4} & \textbf{44.5} & \textbf{44.8} & - & \textbf{38.6} & \textbf{41.5} & \textbf{47.1} & \textbf{44.5} \\
\bottomrule
\end{tabular}}
\caption{Results of adaptation from small to large scale datasets. $^{\dagger}$ is an ideal setting that uses the ground-truth labels of the unlabeled data for class distribution estimation.}
\label{tab: a4-2}
\vskip -0.05in
\end{table}

\vspace{1mm}
\noindent
{\bf Dataset.} As described in \cref{tab:a3}, following the existing DA-OD works, there are four common types of domain shifts in DA-OD. We evaluate our method on these settings and compare it with the state-of-the-arts.

\vspace{1mm}
\noindent
{\bf Network Architecture.} For a fair comparison with the existing DA-OD arts, we switch the backbone from ResNet-50\cite{he2016deep} to VGG-16 \cite{2014Very} and remove the FPN \cite{lin2017feature} neck.

\vspace{1mm}
\noindent
{\bf Implementation Details.} The implementation is nearly the same as SSOD, and more training hyper-parameters can be found in \cref{sec: implementation}. Following previous works, we use $AP_{50}$ as our evaluation metric.

\begin{figure*}[t]
  \centering
  \resizebox{0.98\textwidth}{!}{
  \begin{subfigure}{0.33\linewidth}
    \includegraphics[width=\linewidth]{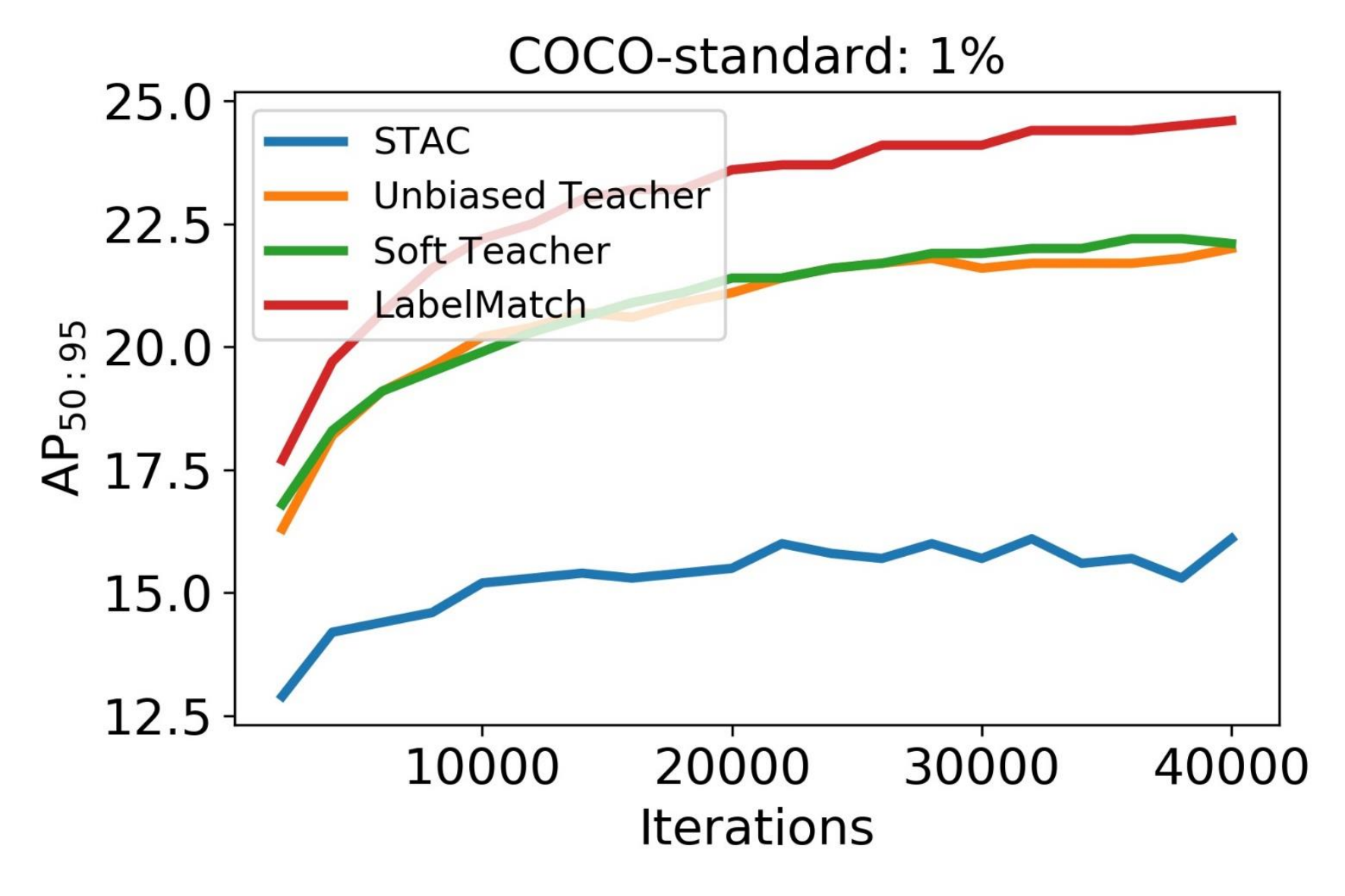}
  \end{subfigure}
  \begin{subfigure}{0.33\linewidth}
    \includegraphics[width=\linewidth]{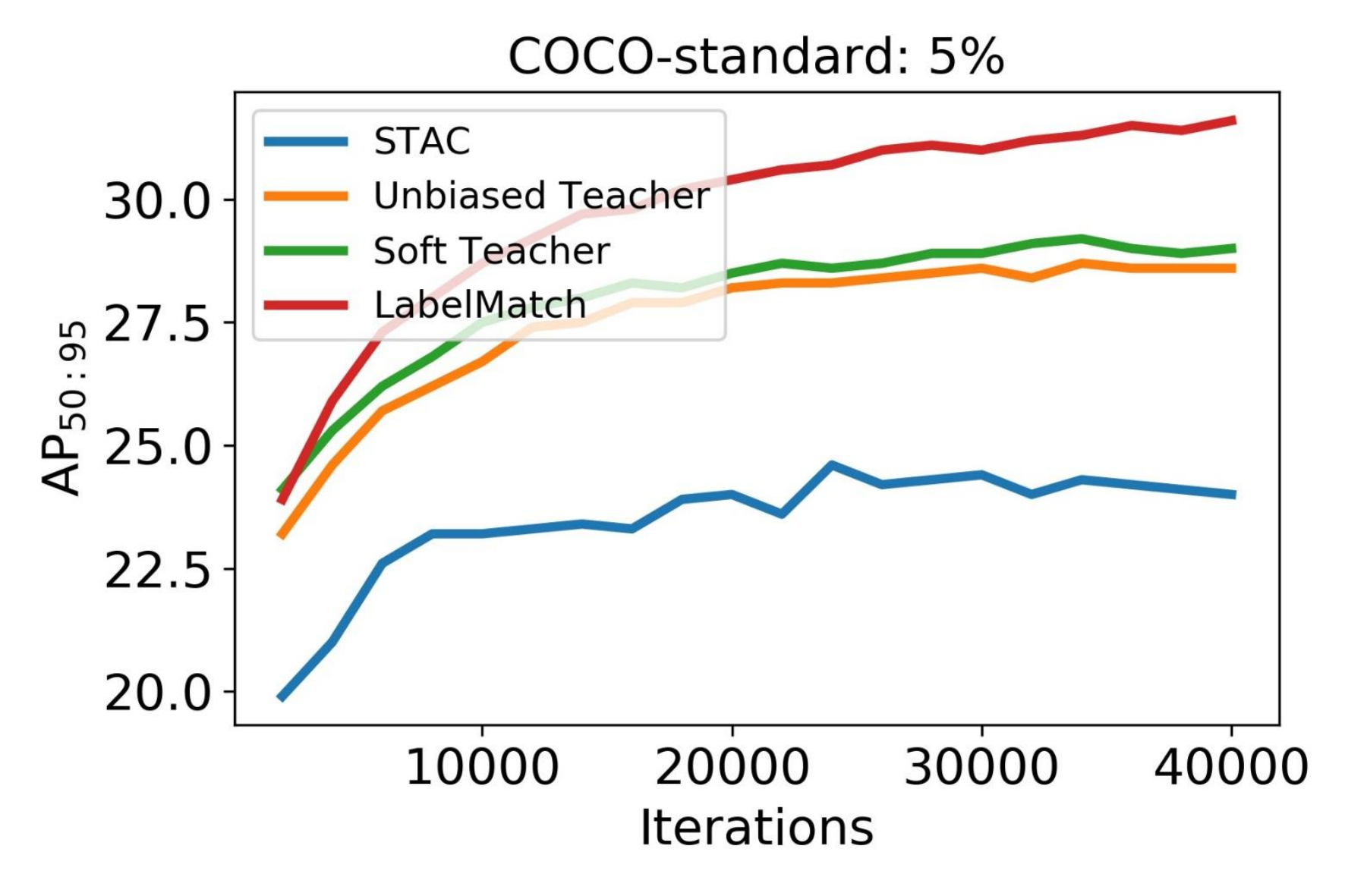}
  \end{subfigure}
  \begin{subfigure}{0.33\linewidth}
    \includegraphics[width=\linewidth]{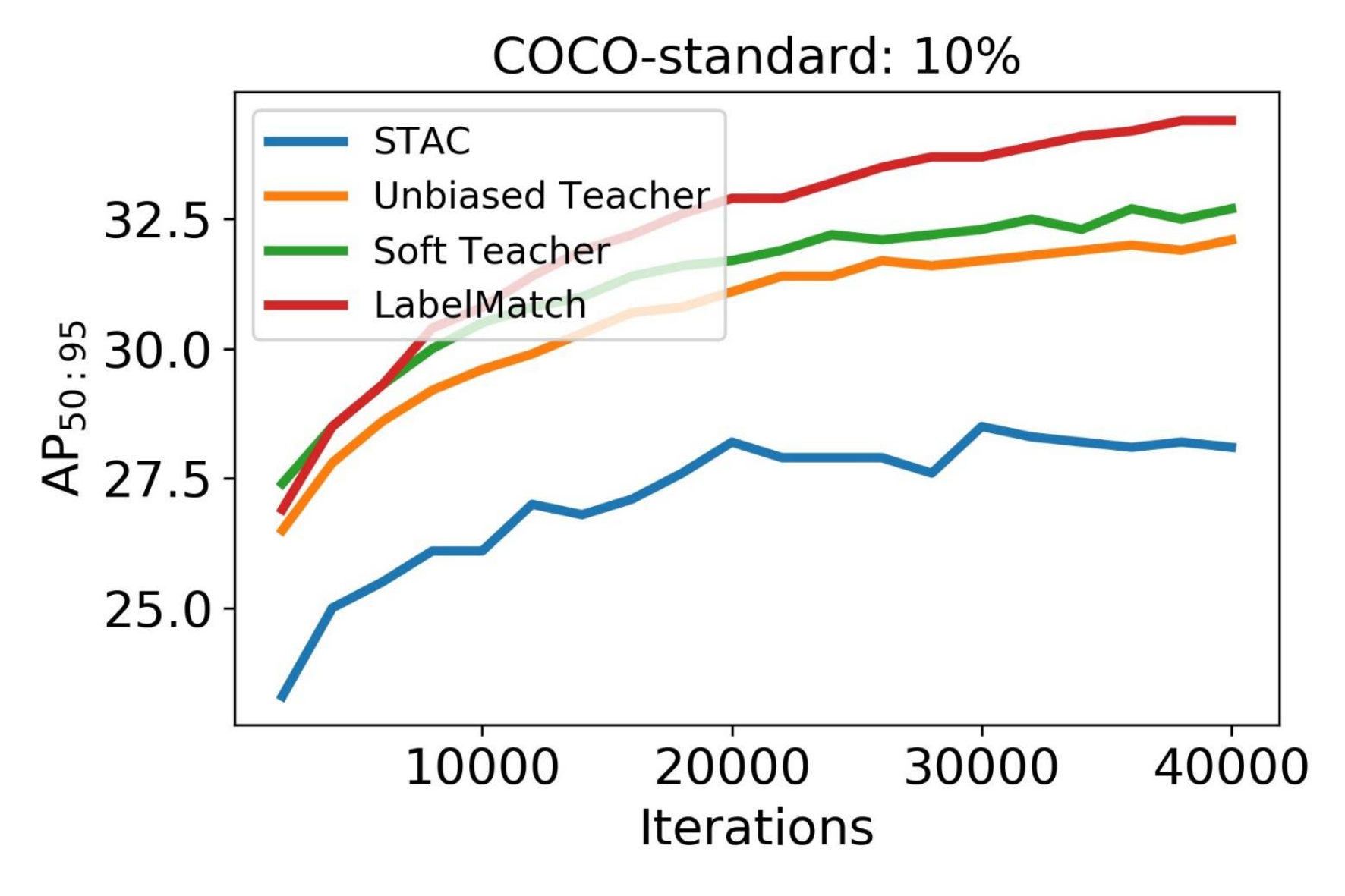}
  \end{subfigure}}
  \caption{Performance ($AP_{50:95}$) comparisons among different state-of-the-art SSOD methods with exactly the same training settings.}
  \label{fig:a5}
  \vskip -0.05in
\end{figure*}

\begin{table}[t]
\parbox{.46\linewidth}{
\centering
\resizebox{0.24\textwidth}{!}{
\begin{tabular}{l|c|c}
\toprule
 Method& $AP_{50}$ & network \\
 \midrule
Source only & 42.2 & FR+VGG \\ \hline
CVPR2019:SW-Faster \cite{saito2019strong} & 37.9 & FR+VGG \\
CVPR2020:GPA \cite{xu2020cross} & 47.9 & FR+R50 \\
CVPR2021:MeGA \cite{vs2021mega} & 43.0 & FR+VGG \\
ICCV2021:SimROD \cite{ramamonjison2021simrod} & 47.5 & YOLOv5 \\ \hline
LabelMatch (Ours) & 51.0 & FR+VGG \\
LabelMatch$^{\dagger}$ (Ours) & \textbf{52.2} & FR+VGG \\
\bottomrule
\end{tabular}}
\caption{Results of adaptation across cameras. FR: Faster-RCNN. $^{\dagger}$ is an ideal setting that uses the ground-truth labels of the unlabeled data for class distribution estimation.}
\label{tab: a5-1}
}
\quad
\parbox{.46\linewidth}{
\centering
\resizebox{0.24\textwidth}{!}{
\begin{tabular}{l|c|c}
\toprule
 Method& $AP_{50}$ & network \\
 \midrule
 Source only & 36.5 & FR+VGG \\ \hline
 CVPR2019:SW-Faster \cite{saito2019strong} & 40.7 & FR+VGG \\
 CVPR2020:GPA \cite{xu2020cross} & 47.6 & FR+R50 \\
 CVPR2021:MeGA \cite{vs2021mega} & 44.8 & FR+VGG \\
 ICCV2021:SimROD \cite{ramamonjison2021simrod} & 52.1 & YOLOv5 \\ \hline
 LabelMatch (Ours) & 52.7 & FR+VGG \\
 LabelMatch$^\dagger$ (Ours) & \textbf{53.8} & FR+VGG \\
\bottomrule
\end{tabular}}
\caption{Results of adaptation from synthetic to real. VGG: VGG-16. $^{\dagger}$ is an ideal setting that uses the ground-truth labels of the unlabeled data for class distribution estimation.}
\label{tab: a5-2}
}
\end{table}

\vspace{1mm}
\noindent
{\bf Results.} To examine the prior dependence on the consistent class distribution assumption, we evaluate LabelMatch in two class distribution estimation manners: 1) The first one is the same as described in the main body of the paper, which estimates the class distribution of the unlabeled target data by the annotations of the labeled source data; 2) The second one is an ideal setting, which determines the class distribution of the unlabeled target data by the ground-truth labels of the unlabeled data.
\begin{itemize}[leftmargin=12pt, topsep=2pt, itemsep=0pt]
\item \underline{Normal$\to$Foggy:} This scenario is different from the following DA-OD settings. In this scenario, the labeled source data and the unlabeled target data meet exactly the same class distribution since the target foggy data is rendered from the normal source data via a foggy translation model. As shown in \cref{tab: a4-1}, benefited from the given class distribution, we achieve a +21.5 mAP improvement over the ``source only'' baseline, exceeding previous state-of-the-arts by a large margin.
\item \underline{Small$\to$Large:} Although there exists bias between the labeled and unlabeled data on foreground-foreground class distribution ($KL=0.36$) and foreground-background ratio (18.5 boxes/img vs. 13.9 boxes/img), our method can still achieve 38.8 mAP, surpassing all the previous arts as far as we know. With access to the accurate class distribution (the ideal setting), our method can be further improved to 44.5 mAP.
\item \underline{Across cameras \& Synthetic$\to$Real:} In these settings, there is only one foreground class and exits foreground-background ratio bias (4.3 boxes/img vs. 9.6 boxes/img and 5.8 boxes/img vs. 9.6 boxes/img). Even using a biased class distribution, our method can still achieve satisfactory results. And our method can get further improvement equipped with the accurate class distribution (aka the ideal setting).
\end{itemize}

These DA-OD experiments demonstrate the robustness of the proposed LabelMatch framework, since the introduction of proposal self-assignment and RPLM weaken the prior dependence on the consistent class distribution assumption. From another perspective, these experiments also indicate that an accurate class distribution estimation can further promote the performance of DA-OD, emphasizing the importance of class distribution estimation. How to estimate an accurate class distribution when the labeled data and the unlabeled data are drawn from two different data distributions is an interesting future work.

\begin{table}[t]
  \centering
  \resizebox{0.48\textwidth}{!}{
  \begin{tabular}{@{}lcccccccc@{}}
    \toprule
    Method & Loss & Threshold & 1\% & 5\% & 10\% \\
    \midrule
    STAC \cite{sohn2020a} & Cross-Entropy & 0.9 & 16.1 & 24.0 & 28.1 \\
    Unbiased Teacher \cite{liu2021unbiased} & Focal-Loss & 0.7 & 22.0 & 28.6 & 32.1 \\
    Soft Teacher$^\star$ \cite{xu2021end} & Cross-Entropy& 0.9 & 22.1 & 29.0 & 32.7 \\
    LabelMatch (Ours) & Cross-Entropy & ACT & \textbf{24.6} & \textbf{31.5} & \textbf{34.6} \\
    \bottomrule
  \end{tabular}}
  \caption{Benchmark results on COCO-standard: our re-implementations with exactly the same training details and data augmentation strategies. $^\star$ denotes the re-implementation without box-jitter trick. We only run 1-fold using the ablation training setting due to the limitation of computation resources. }
  \label{tab:a6}
\end{table}

\begin{table*}[t]
  \centering
    \resizebox{0.98\textwidth}{!}{
    \begin{tabular}{p{0.30\textwidth}>{\centering}p{0.15\textwidth}>{\centering}p{0.15\textwidth}>{\centering}p{0.12\textwidth}>{\centering}p{0.12\textwidth}>{\centering\arraybackslash}p{0.12\textwidth}}
    \toprule
    training setting & COCO-standard & COCO-additional & VOC & Ablation & DA-OD \\
    \midrule
    batch size for labeled data & 16 & 32 & 4 & 32 & 16 \\
    batch size for unlabeled data & 16 & 32 & 4 & 32 & 16 \\
    learning rate & 0.01 & 0.02 & 1.25e-3 & 0.02 & 0.016 \\
    learning rate step & - & (360K, 480K) & - & - & -\\
    iterations & 160K & 540K & 160K & 40K & 20K \\
    unsupervised loss weight $\lambda$ & 2.0 & 2.0 & 2.0 & 2.0 & 2.0 \\
    EMA rate & 0.996 & 0.996 & 0.996 & 0.996 & 0.9996 \\
    reliable ratio $\alpha$ & 0.2 & 0.2 & 0.2 & 0.2 & 0.2 \\
    mean score thresh $T_{score}$ & 0.8 & 0.8 & 0.8 & 0.8 & 0.8 \\
    mean iou thresh $T_{iou}$ & 0.8 & 0.8 & 0.8 & 0.8 & 0.8 \\
    multi-scale (strong augmentation) & (0.2, 1.8) & (0.2, 1.8) & (0.2, 1.8) & (0.5, 1.5) & (0.5, 1.5) \\
    test score thresh & 0.001 & 0.001 & 0.001 & 0.001 & 0.001 \\
    \bottomrule
  \end{tabular}}
  \caption{Training settings for different datasets and different tasks. ``Ablation'' means the training setting of the ablation studies in the main body of the paper, which is also used in all SSOD experiments in the Appendix.}
  \label{tab:a7}
\end{table*}

\begin{table*}[t]
  \centering
    \resizebox{0.98\textwidth}{!}{
    \begin{tabular}{m{0.15\textwidth}m{0.07\textwidth}m{0.31\textwidth}>{\arraybackslash}m{0.50\textwidth}}
    \toprule
    \hline
    \multicolumn{4}{c}{Weak Augmentation} \\
    \midrule
    Process & Prob & Parameters & Descriptions \\ \hline
    Horizontal Flip & 0.5 & None & None \\ \hline
    Multi-Scale & 1.0 & scale=(500, 800) & The short edge of image is random resized from 500 to 800.\\
    \hline\hline
    \multicolumn{4}{c}{Strong Augmentation} \\ \hline
    Process & Prob & Parameters & Descriptions \\ \hline
    Horizontal Flip & 0.5 & None & None \\ \hline
    Multi-Scale & 1.0 & ratio=(0.2, 1.8) & The short edge of image is random resized from $0.5l_{short}$ to $1.5l_{short}$.\\ \hline
    Color Jittering & 0.8 & (brightness, contrast, saturation, hue) = (0.4, 0.4, 0.4, 0.1) & Brightness factor is chosen uniformly form [0.6, 1.4], contrast factor is chosen uniformly from [0.6, 1.4], saturation factor is chosen uniformly from [0.6, 1.4], and hue value is chosen uniformly from [-0.1, 0.1]. \\ \hline
    Grayscale & 0.2 & None & None \\ \hline
    GaussianBlur & 0.5 & (sigma\_x, simga\_y)=(0.1, 2.0) & Gaussian filter with $\sigma_x=0.1$ and $\sigma_y=2.0$ is applied \\ \hline
    CutoutPattern1 & 0.7 & scale=(0.05, 0.2), ratio=(0.3, 3.3) & Randomly selects a rectangle region in an image and erases its pixels. \\ \hline
    CutoutPattern2 & 0.7 & scale=(0.02, 0.2), ratio=(0.1, 6.0) & Randomly selects a rectangle region in an image and erases its pixels. \\ \hline
    CutoutPattern3 & 0.7 & scale=(0.02, 0.2), ratio=(0.05, 8.0) & Randomly selects a rectangle region in an image and erases its pixels. \\ \hline
    \bottomrule
  \end{tabular}
  }
  \caption{Details of data augmentations. In our ablation study, we use multi-scale with ratio=(0.5, 1.5) in order to use large batch size.}
  \label{tab:a8}
\end{table*}

\section{MMDetection-based SSOD Codebase}
\label{sec: ssod}
Since different SSOD algorithms use different data augmentation strategies which have great impact on the performance, we build a unified MMDetection-based SSOD codebase for a fair comparison, named MMDet-SSOD for short, containing STAC \cite{sohn2020a}, Unbiased-Teacher \cite{liu2021unbiased}, Soft-Teacher \cite{xu2021end} and LabelMatch.

We comprehensively run all algorithms in our MMDet-SSOD on COCO-standard dataset using the ablation training setting, and report the performance in \cref{tab:a6} and \cref{fig:a5}. It is worth mentioning that the data augmentation, training iterations, batch size, and other training settings are all kept the same among these algorithms for a fair comparison. The entire source code will be released soon to support the development of SSOD in the community.

\section{Implementation and Training Details}
\label{sec: implementation}

\noindent
{\bf Training.} We utilize different training settings for different datasets in our implementation. We use the SGD optimizer with a momentum rate 0.9 and weight decay 0.0001 in all experiments. The different training settings are summarized in \cref{tab:a7}.

\vspace{3mm}
\noindent
{\bf Data augmentation.} Our data augmentation strategies are modified from Unbiased Teacher \cite{liu2021unbiased}, and the details are shown in \cref{tab:a8}. The weak augmentation is applied to the unlabeled data for pseudo labeling, and the strong augmentation is applied to both labeled and unlabeled data for model training. In our implementation, no cutout augmentation is applied to the labeled data when using strong data augmentation. In order to save computation resources, we use multi-scale with ratio=(0.5, 1.5) in the ablation studies.

\end{document}